\documentclass{article}

\usepackage{PRIMEarxiv}

\usepackage[utf8]{inputenc} 
\usepackage[T1]{fontenc}    
\usepackage{hyperref}       
\usepackage{url}            
\usepackage{booktabs}       
\usepackage{amsfonts}       
\usepackage{nicefrac}       
\usepackage{microtype}      
\usepackage{lipsum}
\usepackage{fancyhdr}       
\usepackage{graphicx}       
\graphicspath{{media/}}     

\title{Ultrafast-and-Ultralight ConvNet-based Intelligent Monitoring System 
for Diagnosing Early-Stage Mpox Anytime and Anywhere}

\author{
  Yubiao Yue \\
  Guangzhou Medical University \\
  \texttt{jiche2020@126.com} \\
   \And
    Xiaoqiang Shi \\
  Chinese Academy of Sciences \\
  \texttt{xiaoqiang.s@outlook.com} \\
   \And
    Li Qin \\
  Chinese Academy of Sciences \\
  \texttt{486236802@qq.com} \\
   \And
    Xinyue Zhang \\
  Guangzhou Medical University \\
  \texttt{moonkkaabc@163.com} \\
   \And
    Jialong Xu \\
  Guangzhou Medical University \\
  \texttt{jialong\_xu18@163.com} \\
   \And
    Zipei Zheng \\
  Guangdong Polytechnic Normal University \\
  \texttt{1158138016@qq.com} \\
   \And
    Zhenzhang Li \\
  Guangzhou Medical University \\
  \texttt{zhenzhangli@gpnu.edu.cn} \\
   \And
    Yang Li \\
  Guangzhou Medical University \\
  \texttt{lychris@sina.com} \\
}

\begin{document}
\maketitle

\begin{abstract}
Due to the absence of more efficient diagnostic tools, the spread of mpox continues to be unchecked. Although related studies have demonstrated the high efficiency of deep learning models in diagnosing mpox, key aspects such as model inference speed and parameter size have always been overlooked. Herein, an ultrafast and ultralight network named Fast-MpoxNet is proposed. Fast-MpoxNet, with only 0.27M parameters, can process input images at 68 frames per second (FPS) on the CPU. To detect subtle image differences and optimize model parameters better, Fast-MpoxNet incorporates an attention-based feature fusion module and a multiple auxiliary losses enhancement strategy. Experimental results indicate that Fast-MpoxNet, utilizing transfer learning and data augmentation, produces 98.40\% classification accuracy for four classes on the mpox dataset. Furthermore, its Recall for early-stage mpox is 93.65\%. Most importantly, an application system named Mpox-AISM V2 is developed, suitable for both personal computers and smartphones. Mpox-AISM V2 can rapidly and accurately diagnose mpox and can be easily deployed in various scenarios to offer the public real-time mpox diagnosis services. This work has the potential to mitigate future mpox outbreaks and pave the way for developing real-time diagnostic tools in the healthcare field.

\end{abstract}


\section{Introduction}
Monkeypox (Mpox), a dreadful zoonotic infectious disease caused by the Monkeypox virus \cite{1} that can be transmitted from animals to humans or from person to person via multiple routes of transmission\cite{2}. Clinically, it usually manifests as skin rash, fever, and lymphadenopathy, and may lead to a range of complications\cite{3}, with a mortality rate of 3-6\% among infected individuals, which can be higher among young children\cite{4}. Like COVID-19, Monkeypox was declared a Public Health Emergency of International Concern (PHEIC) by the World Health Organization, posing a new threat to global public health. Like many infectious diseases, timely, swift and accurate diagnosis of early-stage Monkeypox cases is critical for responding to the spread crisis of Monkeypox virus \cite{5}.

Owing to the paramount importance of recall and specificity in monkeypox diagnosis, the PCR method is recognized as the gold standard in clinical practice\cite{6}. In addition, antibody testing and electron microscopy are also used for Monkeypox diagnosis. Regrettably, despite their high sensitivity and stability, these methods require expensive equipment and materials, skilled technicians, and lengthy reporting times, rendering them unsuitable for widespread and convenient application in various real-world settings. In critical juncture, relying solely on these conventional diagnostic techniques could result in many patients being unable to receive timely diagnoses, leading to the deterioration of their conditions and the rampant spread of the Monkeypox virus. Furthermore, the early clinical manifestations of monkeypox, which are similar to those of influenza and common dermatological diseases, combined with the pre-symptomatic transmission of the virus\cite{7}, can easily lead to misdiagnosis of patients, which in turn can also easily trigger the outbreak of the monkeypox virus. Based on these facts, the development of a low-cost, non-specialized, swift, and accurate diagnostic technology for early-stage monkeypox is of great significance. 

In recent years, deep learning (DL) models, particularly convolutional neural network have demonstrated remarkable performance in computer vision tasks, such as autonomous driving and industrial defect detection \cite{8}. Surprisingly, such technologies have demonstrated performance comparable to or even surpassing medical experts in clinical diagnosis\cite{9}. Furthermore, in contrast to traditional clinical diagnostic techniques, deep learning-based diagnostic technologies can yield significant cost savings in terms of human, material, and financial resources\cite{10}, while obviating the potential for subjective bias that arises from the clinician's expertise. Such technology offers an objective, reliable, and cost-effective alternative for precise and efficient disease detection and diagnosis. Monkeypox disease is primarily characterized by distinct skin rashes. Employing deep learning models and rash images to swiftly differentiate monkeypox skin lesions from other dermatological conditions presents a promising diagnostic approach, which has been corroborated by relevant researches. However, our review indicates that most of the existing literatures primarily emphasizes model validation and performance, often neglecting crucial aspects such as model parameter size and inference speed. From a pragmatic standpoint, the inference speed of a model is pivotal, especially in environments demanding rapid responses, such as international airports and hospitals during the outbreak. In these settings, due to dense population movement, the spread of the virus can be accelerated and then a multitude of individuals from diverse professions necessitate swift and real-time diagnosis. Concurrently, model parameter size dictates the memory and storage requirements of the device, which directly affects the model's adaptability to various platforms. Furthermore, there appears to be a deficiency in the relevant researches concerning these models' efficacy in diagnosing early-stage monkeypox, model interpretability, and the development of practical application. To more effectively address the proliferation of monkeypox cases, it is imperative to prioritize the model's practical utility and foster the development of real-time diagnosis applications with high practicality.

Given the aforementioned facts, we posit that the linchpin to harnessing deep learning model for effective monkeypox outbreaks control rests in the conceptualization of the novel model imbued with the aforementioned attributes, coupled with the development of pertinent applications. Such applications are able to obviate the need for high-cost computational devices and extensive network services and are highly sensitive to early-stage monkeypox, thereby facilitating their broad deployment across diverse common local devices and settings. This would equip us with a robust arsenal to proactively counter potential monkeypox outbreaks in the future. In this study, our contributions can be distilled into the subsequent five key points: (1)To our knowledge, this research represents the pioneering endeavor focusing on the practicality of intelligent monkeypox diagnostic models. Building upon existing literatures, we offered an in-depth discourse on the model's practicality. We posit that achieving an optimal trade-off between diagnostic performance for early-stage monkeypox, inference speed, and parameter size is paramount for a superior monkeypox diagnostic model suited for real-world applications. (2)Due to the preeminence of ShuffleNetV2\cite{11} in terms of inference speed and parameter size, we drew inspiration from it and proposed Fast-MpoxNet. Specifically, to counteract the diagnostic performance constraints imposed by model capacity, an improved attentional feature fusion module and the multi-auxiliary loss strategy \cite{12} were used. Besides, we utilized the DropBlock\cite{13} and GELU to enhance the model's robustness and nonlinear expressiveness. (3)We executed a rigorous five-fold cross-validation on Fast-MpoxNet and 20 other networks, including those adopted in related studies as well as other widely-acknowledged architectures. Our evaluation harnessed the publicly available Mpox Dataset and transfer learning approach. The evaluation metrics included model parameters size, average FPS (The reciprocal of average inference speed), Accuracy, Precision, Recall, Specificity, F1-score. To further measure the model's suitability for the personal computer(PC)-based real-time diagnosis application, we also proposed an innovative metric—the practicality score. This metric provides a comprehensive perspective by integrating Accuracy, Average FPS, Recall and Specificity. (4)We conducted a comprehensive evaluation on images of monkeypox rashes. Specifically, we tested the Recalls of our model on monkeypox images from various stages and distinct parts of the human body, guaranteeing dependable model performance during deployment. Additionally, we utilized the Gradient-weighted Class Activation Mapping (Grad-CAM)\cite{14} to illuminate the underlying decision-making mechanisms of the model. (5)Based on our research\cite{15}, we developed intelligent monkeypox diagnostic applications suitable for both PC and mobile platforms, named Mpox-AISM V2. PC Mpox-AISM V2 and mobile Mpox-AISM V2 are tailored for medical professionals and the general populace, respectively. Notably, these applications obviate the need for internet connectivity and high-end computational hardware. In the PC application, we also created a facial and hand detector using OpenCV to automatically detect the key parts of human body for further reducing the workload of medical staff. Mpox-AISM V2 excels in accurate, rapid and real-time diagnosis for early-stage monkeypox and can be applied in various real-world settings.

\section{Related Work}
\label{sec:headings}
Despite its identification in the 1960s, monkeypox remained relatively obscure to many, including medical professionals, until the recent outbreak. Consequently, there exists a limited body of research employing deep learning (DL) models for diagnosing monkeypox. Before embarking on this study, we conducted a comprehensive search on the Web of Science using the terms "monkeypox" and "deep learning", and then found 13 related works. In the ensuing section, we will delve into a meticulous examination of the research outcomes and limitations of these published works.

Thieme et al. assembled a large dataset of skin lesion images, consisting of 139,198 images, which included 676 monkeypox images [16] . They then employed a convolutional neural network (CNN) named MPXV-CNN, based on Resnet34, and extensively validated and tested it on this large dataset. MPXV-CNN achieved a recall of 83.00 and 91.00, and a specificity of 96.50 and 89.90 on the validation and test sets, respectively. Finally, Thieme et al developed a web-based application based on this.

In the study of Jaradat et al., they used 117 images they collected themselves, including 45 monkeypox images \cite{17}. They then validated these images using VGG19, VGG16, ResNet50, MobileNetV2, and EfficientNetB3 on augmented data, and found that MobileNetV2 performed the best. MobileNetV2 achieved an accuracy of 98.16\%, and a precision, recall, f1-score, and specificity of 99.00\%, 96.00\%, 98.16\%, and 98.80\%, respectively.

The data used in the study of Abdlhamid et al. consisted of 279 monkeypox and 293 normal skin images from the Monkeypox skin images dataset (MSID) \cite{18}. Their research aim was to achieve better performance in binary classification tasks, for which they proposed two algorithms to improve the accuracy of individual models. The first algorithm, called Binary PSOBER, aimed to select the optimal feature set that could improve classification accuracy. The second algorithm, called SCBER, aimed to optimize the parameters of the neural network. Finally, they combined these algorithms with GoogleNet using transfer learning, achieving an average accuracy of 98.80\% on the adopted dataset after augmentation.

Akin et al. also used 279 monkeypox and 293 normal skin images from the Monkeypox skin images dataset (MSID) \cite{19}. They employed 12 different CNN models for the binary classification of monkeypox and normal skin. They found that the MobileNet V2 model performed the best with an accuracy of 98.25\%, sensitivity of 96.55\%, specificity of 100.00\%, and F1-Score of 98.25\%. Finally, they developed a decision support system based on explainable artificial intelligence (AI) assisted convolutional neural networks (CNNs).

Uzun Ozsahin et al. integrated publicly available datasets with their own collected data to form a dataset of 342 images, including 102 monkeypox and 240 chickenpox images \cite{20}. Their study aimed to perform the binary classification of monkeypox and chickenpox using deep learning. To achieve this, they first enhanced the data and proposed a CNN network that they designed themselves. After testing, their proposed model achieved 99.60\% accuracy on the test set of the adopted dataset.

Alcalá-Rmz et al. utilized the publicly available dataset called Monkeypox Skin Lesion Dataset (MSLD) \cite{21}. The dataset was divided into a training set and a validation set in an 8:2 ratio. They then evaluated the dataset using a model called MiniGoogleNet. The results showed that at the 50th epoch, MiniGoogleNet performed the best, achieving an overall accuracy of 97.08\% on the validation set.

Ahsan et al. used their self-constructed binary dataset containing 76 images and the MSID dataset for their study \cite{22}. They developed a monkeypox diagnostic model for binary and multi-class classification using a transfer learning approach based on generalization and regularization (GRA-TLA). The proposed method was tested on ten different convolutional neural network (CNN) models in three independent studies. Their proposed method combined with Extreme Inception (Xception) achieved accuracies ranging from 77.00\% to 88.00\% in studies one and two, while the residual network ResNet101 demonstrated the best multi-class classification performance in study three, with accuracies ranging from 84.00\% to 99.00\%.

Sahin et al.  utilized the MSLD dataset, and their main objective was to develop a mobile application for binary classification of monkeypox \cite{23}. To achieve this, they validated six lightweight networks using the dataset and transfer learning techniques, and found that MobileNetV2 performed the best. MobileNetV2 achieved an accuracy of 91.11\% and an average inference time of 197ms, 91ms, and 138ms on three mobile devices.

Eid et al. proposed a novel method for accurately predicting diagnosed cases of monkeypox using an optimized long short-term memory (LSTM) deep network, and optimized the hyperparameters of the LSTM-based deep network using the AI-Biruni Earth Radius optimization algorithm \cite{24}. In their experiments, the authors used a public dataset and six different machine learning models, and compared BER-LSTM with four different optimization algorithms, and the experimental results showed that BER-LSTM was effective.

Bala et al. created the MSID dataset, consisting of 770 images, including 279 images of monkeypox \cite{25}. In their paper, the MSID dataset was expanded to 8689 images by using data augmentation techniques. Then, they extensively cross-validated a deep learning-based improved DenseNet-201 CNN model, called MonkeyNet, using both the original and augmented datasets. Experimental results showed that MonkeyNet achieved an accuracy of 93.19\% and 98.91\% on the original and augmented datasets, respectively.

Altun et al.  used a custom dataset and the MSLD dataset in their experiment \cite{26}. They adopted transfer learning strategy on six classic CNN networks and studied the hyperparameters and optimization methods. Then, they established the hyperparameters and trained and evaluated the six models using the custom dataset and MSLD dataset. The results indicated that MobileNetV3-s was the state-of-the-art model, achieving 96.00\%, 97.00\%, 98.00\%, and 99.00\% in terms of Accuracy, recall, F1-score, and AUC, respectively.

Sitaula and Shahi employed two publicly available datasets that were augmented, consisting of 1,754 images with 587 of them containing monkeypox \cite{27}. In this paper, the authors utilized 13 pre-trained CNN models, which were fine-tuned uniformly, and then evaluated. Finally, they employed an ensemble model approach based on a multiple voting mechanism, and the results showed that the best performance was achieved by combining the Xception and DenseNet-169 models, with Precision, Recall, F1-score, and Accuracy reaching 85.44\%, 85.47\%, 85.40\%, and 87.13\%, respectively.

Yasmin et al. utilized the augmented dataset (1428 Monkeypox images  and 1764 other images) from MSLD \cite{28}. They trained and tested six models on the dataset and found the performance of InceptionV3 is best. They further fine-tuned InceptionV3 and proposed an improved network, called PoxNet22. PoxNet22 achieved perfect scores of 100\% in accuracy, precision, and recall metrics.

It is imperative to highlight that while prior researches have garnered commendable outcomes across diverse model evaluation metrics, the absence of a standardized official dataset, coupled with the variability in the number and taxonomy of datasets employed across studies, impedes objective appraisals of the methodologies adopted in these works. Therefore, from the practicality of each work, we summarized the following limitations of previous researches: \textbf{(1)Single Model Metric-Centric Focus:} Certain studies have predominantly concentrated on model validation and performance, sidelining pivotal considerations for real-world applications such as inference speed and model parameter size. \textbf{(2)Lack of Comprehensive Evaluation:} While some research endeavors developed practical applications, they often fall short in providing essential model interpretability and a nuanced evaluation of monkeypox images. Practical deployments necessitate the assessment of model diagnosis ability across diverse bodily regions and progression stages of monkeypox, with an emphasis on facial and hand regions as well as early-stage\cite{16}. \textbf{(3)Limitations for Cloud Server and Heavyweight Models:} The applications in related researches have a certain degree of practicality, but are tethered to cloud servers and employed heavyweight models. Consequently, these applications are susceptible to network fluctuations and the responsiveness of them is constrained by the computational prowess of the cloud servers and the architecture of network. Moreover, these applications are not tailored for PC platforms and cloud servers accrue elevated maintenance overheads, making them less feasible for deployment in high-risk environments like airports, subway stations, and hospitals. Paradoxically, these are the very locales where swift diagnostic solutions are paramount to curb viral transmission. Furthermore, the least Developed Countries often need to grapple with higher incidence rates, but are bereft of internet access \cite{29}. The public there are in dire need of diagnostic tools that are versatile and not internet-dependent. Additionally, cloud-based applications often falter in safeguarding patient confidentiality, potentially engendering patient apprehensions.

In essence, an optimal monkeypox diagnostic application should encapsulate attributes such as offline functionality, high accuracy, robust interpretability, smaller model parameter size, and swift response speed. Table 1 elucidates the advancements and concomitant limitations of each study. Within Table 1, the average FPS signifies the number of images a model can process within a second. In the context of our study, a higher average FPS denotes the model's efficacy in the populous settings. The model parameters size represents the computational memory  during operation and sto Settingsrage space. Typically, a diminutive model parameter size augments the model's applicability across a spectrum of computational devices, inclusive of micro-embedded systems.

\begin{table}[]
\caption{limitations in the related researches.}
\resizebox{1\textwidth}{!}{
\begin{tabular}{@{}llllllll@{}}
\toprule
paper & \begin{tabular}[c]{@{}l@{}}Model\\Name\end{tabular}                        & \begin{tabular}[c]{@{}l@{}}Average\\FPS\end{tabular} & \begin{tabular}[c]{@{}l@{}}ParSize\\(num\_classes=1000)\end{tabular} & \begin{tabular}[c]{@{}l@{}}Model\\Accuracy\end{tabular}  & \begin{tabular}[c]{@{}l@{}}Graded\\evaluation\end{tabular} & Interpretability& \begin{tabular}[c]{@{}l@{}}Application\\type\end{tabular} \\ \midrule
16    & MPXV-CNN(Based   on ResNet34)                                                     & About20                                                     & 21.8M                                                                          & 90.40\%                                                         & Yes                                                               & Yes                                                     & Mobile   Phone(Cloud)                                            \\
17    & MobileNetV2                                                                       & 22                                                          & 3.5M                                                                           & 98.16\%                                                         & No                                                                & Yes                                                     & No                                                               \\
18    & GoogLeNet                                                                         & 22                                                          & 13.0M                                                                          & 98.80\%                                                         & No                                                                & No                                                      & No                                                               \\
19    & MobileNetV2                                                                       & 22                                                          & 3.5M                                                                           & 98.25\%                                                         & No                                                                & Yes                                                     & No                                                               \\
20    & CNN                                                                               & Unknown                                                     & Unknown                                                                        & 99.60\%                                                         & No                                                                & No                                                      & No                                                               \\
21    & MiniGoogLeNet                                                                     & 41                                                          & 1.6M                                                                           & 97.08\%                                                         & No                                                                & No                                                      & No                                                               \\
22    & \begin{tabular}[c]{@{}l@{}}Xception\\ResNet101\end{tabular}                & \begin{tabular}[c]{@{}l@{}}23\\7\end{tabular}        & \begin{tabular}[c]{@{}l@{}}22.0M\\44.5M\end{tabular}                    & \begin{tabular}[c]{@{}l@{}}94.00\%\\99.00\%\end{tabular} & \begin{tabular}[c]{@{}l@{}}No\\No\end{tabular}             & \begin{tabular}[c]{@{}l@{}}Yes\\Yes\end{tabular} & \begin{tabular}[c]{@{}l@{}}No\\No\end{tabular}            \\
23    & MobileNetV2                                                                       & 22                                                          & 3.5M                                                                           & 91.11\%                                                         & No                                                                & Yes                                                     & No                                                               \\
24    & CNN+LSTM                                                                          & Unknown                                                     & Unknown                                                                        & Unknown                                                         & No                                                                & No                                                      & No                                                               \\
25    & \begin{tabular}[c]{@{}l@{}}MpoxNet\\(Based on DenseNet201)\end{tabular}  & About6                                                      & 20.0M                                                                          & 98.91\%                                                         & No                                                                & Yes                                                     & No                                                               \\
26    & MobileNet\_V3\_small                                                              & 60                                                          & 2.5M                                                                           & 96.80\%                                                         & No                                                                & No                                                      & Mobile Phone(Local)                                            \\
27    & Xception combing with DenseNet-169                                              & About   7                                                   & 36.1M                                                                          & 87.13\%                                                         & No                                                                & Yes                                                     & No                                                               \\
28    & \begin{tabular}[c]{@{}l@{}}PoxNet22\\(Based on InceptionV3)\end{tabular} & About   15                                                  & 27.2M                                                                          & 100\%                                                           & No                                                                & No                                                      & No                                                               \\ \bottomrule
\end{tabular}}
\end{table}

\section{Materials and Methods}
\subsection{Data Source}
In this study, one of our main tasks was to classify monkeypox and other skin diseases similar to monkeypox as well as normal human skin. Therefore, two datasets were used: Monkeypox Skin Image Dataset (MSID) \cite{25} and Monkeypox Skin Lesion Dataset (MSLD) \cite{30,31}. The MSID dataset includes four types of images: monkeypox, chickenpox, measles, and normal skin. The MSLD dataset contains two types of images: monkeypox and other skin diseases. Due to the scarcity of monkeypox images, we merged the monkeypox images from MSLD into MSID, resulting in four categories of data: monkeypox, chickenpox, measles, and normal skin. The specific numbers for each category in the MSID and MSLD datasets are shown in Table 2. We refer to the merged dataset of MSID and MSLD as the Mpox Dataset (Figure 1).

\begin{figure}[ht]
    \centering
    \includegraphics[width=1\textwidth]{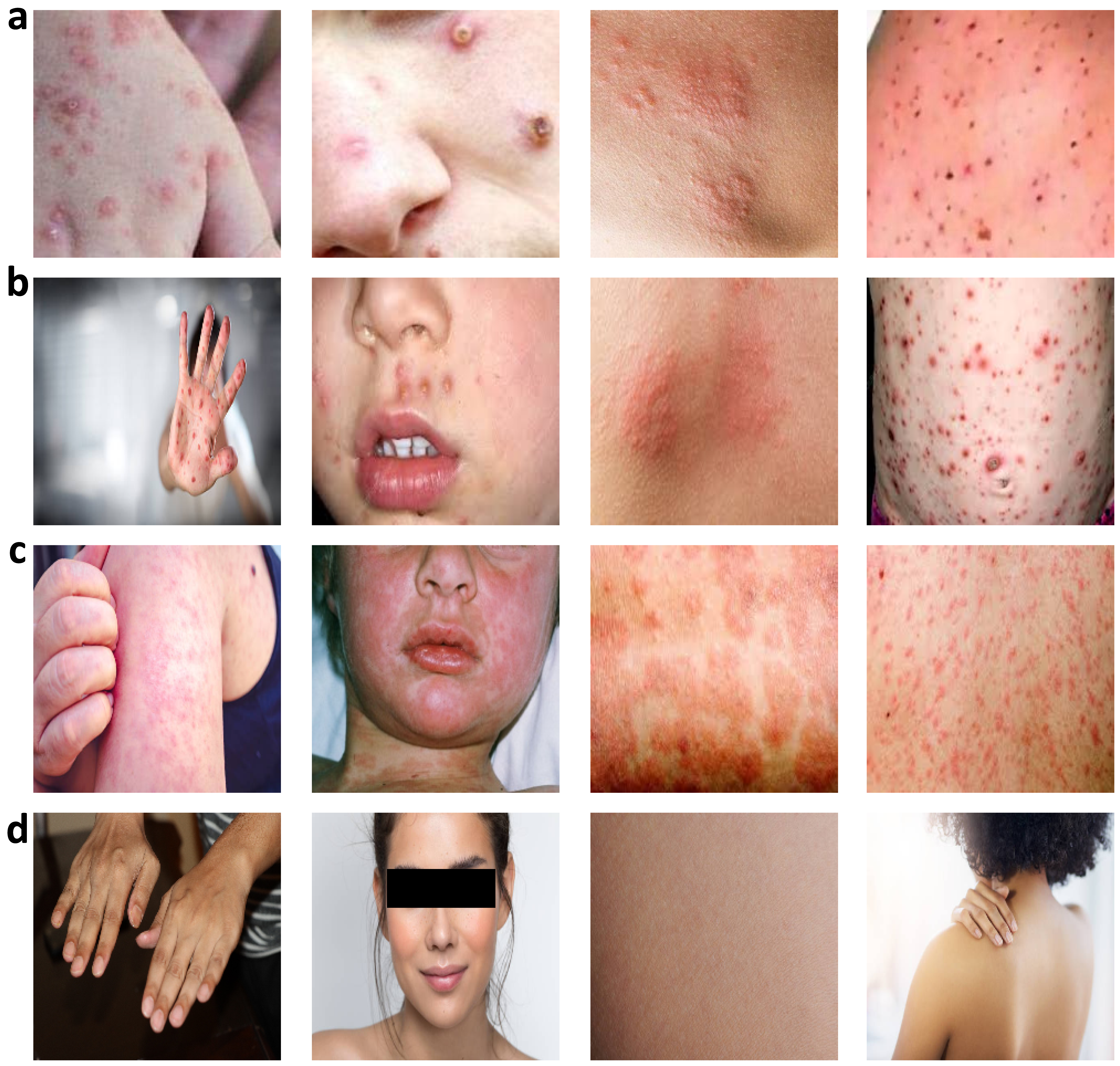}
    \caption{a: Monkeypox samples, b: Chickenpox samples, c: Measles samples, d: Normal samples.}
    \label{fig:enter-label}
\end{figure}

\begin{table}[]
\centering
\caption{Distribution of images in different datasets.}
\resizebox{0.8\textwidth}{!}{
\begin{tabular}{@{}cccccc@{}}
\toprule
\textbf{Dataset\_name} & \textbf{Monkeypox} & \textbf{Measles} & \textbf{Chickenpox} & \textbf{Normal} & \textbf{Others} \\ \midrule
\textbf{MSID}          & 279                & 91               & 107                 & 293             & 0               \\
\textbf{MSLD}          & 102                & 0                & 0                   & 0               & 126             \\
\textbf{Mpox Dataset}  & 381                & 91               & 107                 & 293             & 0               \\ \bottomrule
\end{tabular}}
\end{table}

\subsection{Data augmentation}
Data augmentation is a crucial technique employed during the training phase of deep learning models, particularly in the domain of computer vision \cite{32}. It serves to artificially expand the available training dataset by generating new, yet diverse samples derived from the original data through various transformations, such as rotation, scaling, flipping, and cropping. The primary objective of data augmentation is to enhance the model's ability to generalize and reduce overfitting, thus improving its performance on previously unseen data.

\begin{table}[]
\caption{The details of data augmentation adopted in our work.}
\resizebox{1\textwidth}{!}{
\begin{tabular}{@{}llll@{}}
\toprule
Strategy name for data augmentation & Parameters value                       & Action                                                                                                                                                                                                                                  &  \\ \midrule
Fliplr                              & 0.5                                    & Flip the input image horizontally with a 50\% chance.                                                                                                                                                                                   &  \\
Flipud                              & 0.5                                    & Flip the input image vertically with a 50\% chance.                                                                                                                                                                                     &  \\
Crop                                & 0$\sim$0.2                             & Randomly Crop the input image inma ratio of 0\% to 20\% of the input image.                                                                                                                                                             &  \\
GaussianBlur                        & Sigma: 0$\sim$0.5                      & Blur the input image with a gaussian kernel with a sigma of 0 to 0.5.                                                                                                                                                                   &  \\
LinearContrast                      & 0.5$\sim$1.5                           & \begin{tabular}[c]{@{}l@{}}Modify the contrast of the input image according to 127 + alpha*(v-127)`, \\ where v is a pixel value and alpha is sampled uniformly from the interval {[}0.5, 1.5{]}.\end{tabular}                          &  \\
AdditiveGaussianNoise               & Scale: 0$\sim$0.05*255                 & \begin{tabular}[c]{@{}l@{}}Add gaussian noise to the input image, sampled once per pixel from \\ a normal distribution N(0, s), where s is sampled per image and varies between 0 and 0.2*255.\end{tabular}                             &  \\
AddToSatutration                    & -50$\sim$50                            & \begin{tabular}[c]{@{}l@{}}Sample random values from the discrete uniform range {[}-50, 50{]}, \\ and add them to the saturation of the input image, \\ i.e. to the S channel in HSV color space.\end{tabular}                          &  \\
MultiplyBrightness                  & 0.5$\sim$1.5                           & \begin{tabular}[c]{@{}l@{}}Convert the input image to a color space with a brightness-related channel, extract that channel, multiply \\ it by a factor between 0.5 and 1.5, and convert back to the original color space.\end{tabular} &  \\
Affine(Scale)                       & "x": 0.5$\sim$1.5, “y”: 0.5$\sim$1.5   & \begin{tabular}[c]{@{}l@{}}Scale the input image to a value of 50 to 150\% of their original size, \\ but do this independently per axis (i.e. sample two values per image).\end{tabular}                                               &  \\
Affine(translate)                   & "x": -0.2$\sim$0.2, “y”: -0.2$\sim$0.2 & Translate the input image by -20 to +20\% on x- and y-axis independently.                                                                                                                                                               &  \\
Affine(Shear)                       & -16$\sim$16                            & Shear the input image by -16 to 16 degrees.                                                                                                                                                                                             &  \\
Affine(Rotate)                      & -45$\sim$45                            & Rotate the input image by -45 to 45 degrees.                                                                                                                                                                                            &  \\ \bottomrule
\end{tabular}}
\end{table}

Incorporating data augmentation into the training pipeline enables the model to better cope with variations in the input data, thereby fostering a more robust and reliable model. This is particularly important when working with limited or imbalanced datasets, as data augmentation helps alleviate the shortage of samples and balance the class distribution \cite{33}, resulting in a more effective learning process. Moreover, data augmentation facilitates the discovery of underlying patterns and invariant features that are pivotal for accurate classification, especially in complex real-world settings.
\begin{figure}[h]
    \centering
    \includegraphics[width=1\textwidth]{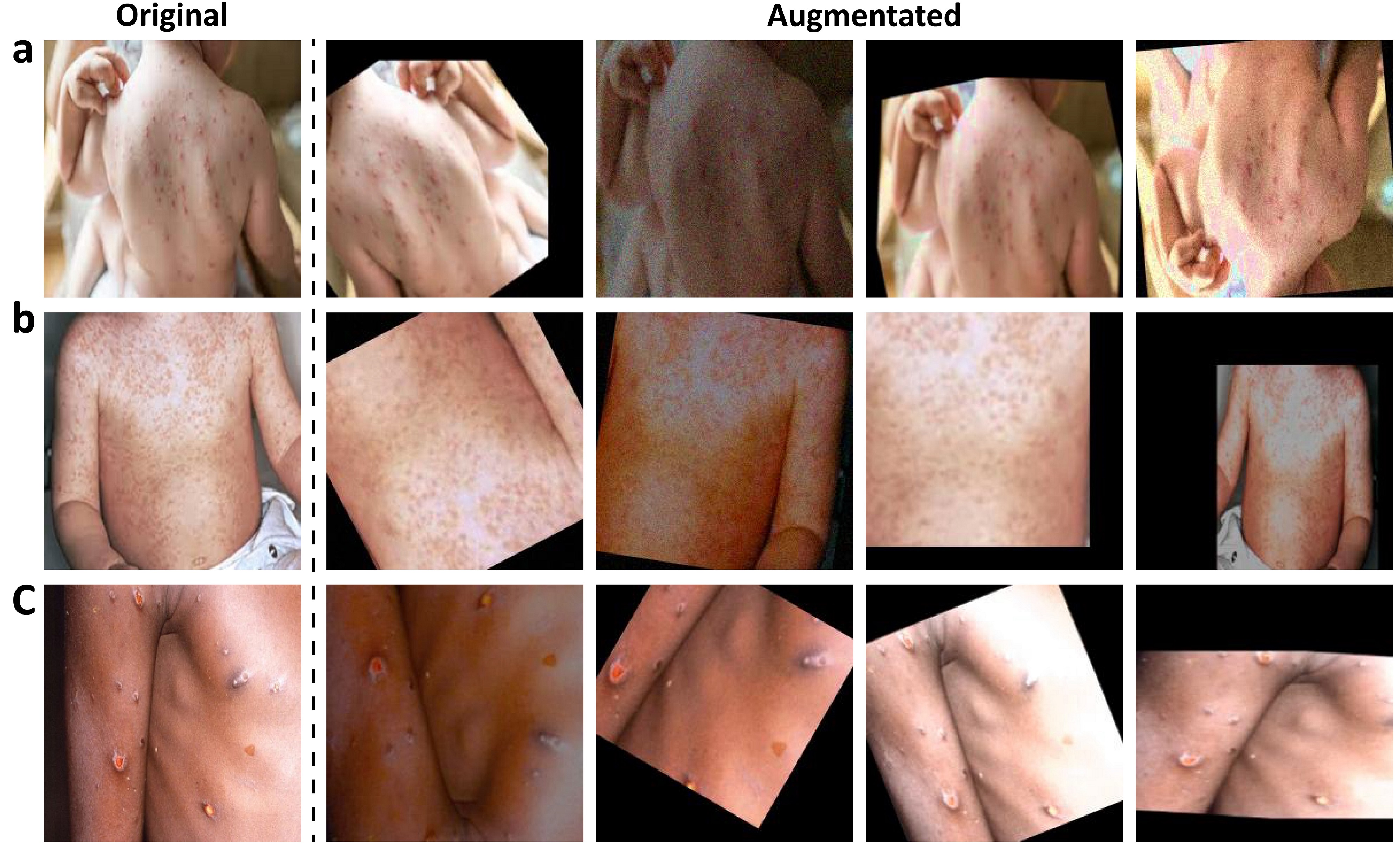}
    \caption{Some samples of augmented dataset and comparison between original images and augmented images. a: Monkeypox samples, b: Chickenpox samples, c: Measles samples.}
    \label{fig2}
\end{figure}
Here, taking into account practical application needs, we considered changes in image clarity, angle, shooting distance, color, and noise when augmenting Mpox Dataset. We then employed a total of 12 augmentation strategies, and Table 3 provides detailed descriptions of each strategy. In practice, we randomly selected 6 augmentation strategies from the 12 augmentation strategies, and applied them to each original image that required augmentation. Additionally, the application of these 6 augmentation strategies was also done in a random order.

In this work, we performed data augmentation on the Chickenpox, Measles, and Monkeypox images in the Mpox Dataset, with the original number of images for these three categories being 107, 91, and 381, respectively. After augmentation, the number of images for each category increased to 2354, 2366, and 2667. For the Normal category, we used images from Hands and palm images dataset \cite{34} and SFA IMAGE DATABASE \cite{35}  to augment the data, resulting in a significant increase in diversity. The number of Normal images increased from 279 to 2328. In Figure 2, we present some examples of augmented Chickenpox, Measles, and Monkeypox images. We abbreviated augmented Mpox Dataset as Aug Dataset.

\subsection{Evalutaion Metrics}
Accuracy, Precision, Recall, Specificity, and F1-score are commonly used evaluation metrics in deep learning. In Equation (1) to Equation (5), Accuracy measures the proportion of correct classifications, while Precision represents the proportion of true positives out of all predicted positives. Recall measures the proportion of true positives out of all actual positives. Specificity measures the proportion of true negatives out of all actual negatives. F1-score, which measures the classifier's overall performance, is the harmonic mean of Precision and Recall. These metrics are commonly used to assess the performance of classifiers and help determine the optimal model for a given task. In these equations, TP, TN, FP and FN are the numbers of true positives, true negatives, false positives and false negatives.

In certain studies, floating-point operations (FLOPs) are utilized to measure the inference speed of a model. Theoretically, a lower FLOPs suggests lower computational demands, implying a faster inference speed. However, in real-world hardware environments, FLOPs may not provide an accurate reference of model speed due to the presence of various optimization and computational operations. Put succinctly, there isn't a consistent positive correlation between FLOPs and actual inference speed in practical settings. Given the objectives of our research, we used Average Frame per Second (Average FPS) as the primary metric to evaluate the model's inference speed. Contrasted with FLOPs, Average FPS offers a direct insight into the model's inference speed on a designated platform. Specifically, a higher Average FPS indicates that the system can swiftly process input via real-time video, ensuring a seamless user experience and enabling the model to screen a larger populace in areas with high footfall. This metric is delineated in Equation (6). In Equation (6), N symbolizes the quantity of images designated for testing; a larger N value enhances the reliability of the outcome. In addition,  and   respectively represent the end time and start time when the network inferences the image. The result of model parameter size is calculated by the Pytorch framework.

\begin{equation}
Accuracy = \frac{TP + TN}{TP + TN + FP + FN}\nonumber
\end{equation}

\begin{equation}
Precision = \frac{TP}{TP + FP}
\end{equation}

\begin{equation}
Recall = \frac{TP}{TP + FN}
\end{equation}

\begin{equation}
Specificity = \frac{TN}{TN + FP}
\end{equation}

\begin{equation}
F1\_score = \frac{2 \times Precision \times Recall}{Precision + Recall}
\end{equation}

\begin{equation}
AverageFPS = \left\lceil \frac{1}{N} \sum_{i = 1}^N \frac{1}{t_i^{out} - t_i^{in}} \right\rceil
\end{equation}

In diverse research works, researchers often find themselves evaluating a plethora of models, necessitating a comprehensive evaluation of various metrics to discern the model most apt for their specific task. The numerous results of these metrics may obfuscate a clear, direct assessment of a model's merits and demerits. In the context of this study, our primary focus is the development of  PC-based application. This emphasis stems from the rationale that medical staff can utilize PC-based application for proactively screening the public in high-risk settings via real-time video, rather than relying on patients to autonomously utilize mobile applications for self-diagnosis. Given the unique attributes of medical imagery and settings for PC Applications, we introduced a novel metric termed the "Practicality Score." This metric is predicated upon four pivotal metrics: Accuracy, Average FPS, Recall, and Specificity. It first uses linear normalization to ascertain an individual model's score for a given metric, followed by a weighted summation to compute the overall Practicality Score. Such a metric offers a more lucid and intuitive representation of a model's strengths and weaknesses in real-time diagnosis application. The calculation equations for Practicality Score are from Equation(7) to Equation (11). In these equations, w1, w2, w3 and w4 correspond to the weights of each metric, respectively and the variables indexed by i correspond to a specific model. Besides, the variables indexed by max and min respectively correspond to the maximum and minimum values of a certain metric across all models. Here, w1, w2, w3 and w4 were set to 0.15, 0.4, 0.35, and 0.15, respectively, according to the importance of different metrics.
\begin{equation}
Accuracy' = \frac{Accuracy_i - Accuracy{y_{min}}}{Accuracy{y_{max}} - Accuracy{y_{min}}}
\end{equation}

\begin{equation}
AverageFPS' = \frac{{FP{S_i} - FP{S_{min}}}}{{FP{S_{max}} - FP{S_{min}}}}    
\end{equation}

\begin{equation}
Recall' = \frac{{Recal{l_i} - Recal{l_{min}}}}{{Recal{l_{max}} - Recal{l_{min}}}}  
\end{equation}

\begin{equation}
Specificity' = \frac{{Specificit{y_i} - Specificit{y_{min}}}}{{Specificit{y_{max}} - Specificit{y_{min}}}}
\end{equation}

\begin{equation}
Practicality\_score = {w_1} \times Accuracy' + {w_2} \times FPS' + {w_3} \times Recall' + {w_4} \times Specificity'    
\end{equation}

\subsection{Transfer Learning}
Transfer learning is a popular technique in deep learning that aims to improve the performance of a machine learning model by leveraging the knowledge learned from a pre-trained model \cite{36}. It involves using a pre-trained model as a starting point and adapting it for a new task by fine-tuning some or all of its parameters. This approach can be particularly effective when the available labeled data for the new task is limited or when training a model from scratch is computationally expensive. Transfer learning has been applied successfully in various applications, including computer vision, natural language processing, and speech recognition.

In this work, we employed transfer learning with pre-trained models on the ImageNet dataset. Specifically, we used the pre-trained models as a starting point and fine-tuned their weights on our target task. The ImageNet pre-training provides a strong initialization for the model, allowing it to learn better feature representations and achieve better performance with less labeled data for Monkeypox diagnosis task.

\subsection{The overall network structure of Fast-MpoxNet}
Here, we chose the ShuffleNetV2\_X0\_5 (Short for ShuffNetV2) as our baseline via comparing the average FPS and model parameter sizes of various networks. Inspired by Attentional Feature Fusion [12], we improved it and proposed Attention-Based Local and Global Fusion Module (ABLGFM) as the feature fusion module, and then applied it to ShuffleNetV2 to design Fast-MpoxNet. 

The overall network architecture of Fast-MpoxNet is illustrated in Figure 3, which consists of three main components: stem, backbone(stage2~5), auxiliary classification heads, and feature fusion module. The stem and stage2~5 of Fast-MpoxNet have the same internal structure as ShuffleNetV2. It mainly consists of a large number of depth-wise separable convolutions. Each input image to Fast-MpoxNet undergoes an image pre-processing module, where the image is first randomly scaled and cropped to 224×224, and then normalized using mean [0.485, 0.456, 0.406] and standard deviation [0.229, 0.224, 0.225]. Subsequently, the stem and backbone of Fast-MpoxNet down-samples the image and extracts its features at different levels. Specifically, in the stem blcok, the image is first down-sampled by a convolution layer with stride 2 and kernel size 3, followed by a max pooling layer with stride 2, resulting in a 4× down-sampling of the image. Then, Stage 2 to Stage 5 down-sample the feature maps from stem block.
\begin{figure}[h]
    \centering
    \includegraphics[width=1\textwidth]{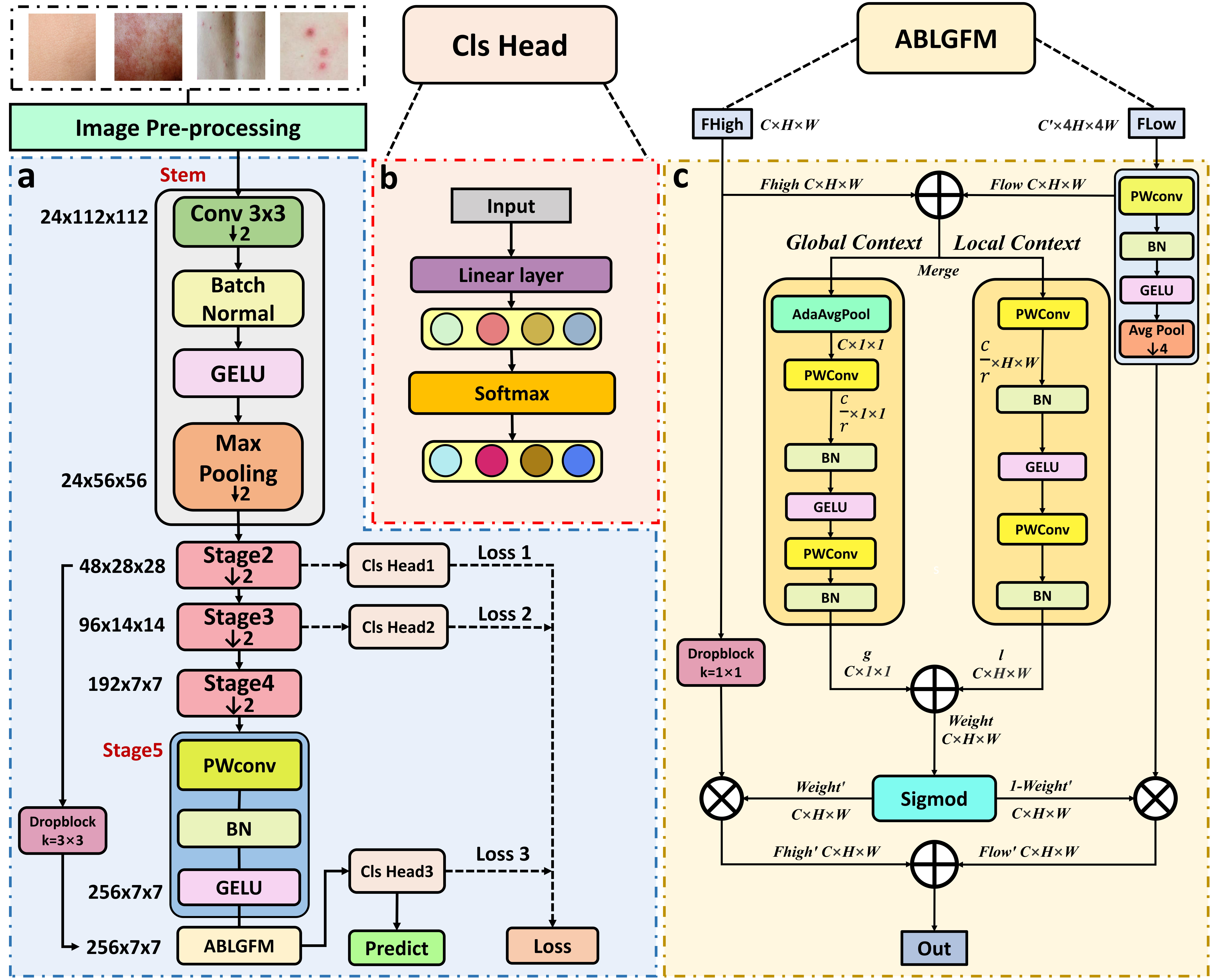}
    \caption{a: The overall architecture of Fast-MpoxNet. Here, PWConv is Point-wise Convolution which can change the number of channels of the feature map and has the characteristics of low computational power consumption and low memory access frequency. BN represents Batch Normalization. Besides, C×H×W represents the number of channels, image height, and image width. b: Internal structure of Cls Head. c: Internal structure of Attention-Based Local and Global Fusion Module (ABLGFM).}
    \label{fig3}
\end{figure}
Furthermore, we added auxiliary classification heads, namely Cls Head1 and Cls Head2, to the semantic branches of Stage 2 and Stage 3. The outputs of Stage 2 and Stage 3 are fed into Cls Head1 and Cls Head2, respectively, for auxiliary loss calculation. Finally, we used ABLGFM to fuse the extracted features from Stage 2 and Stage 5, and feed the output into Cls Head3 for image classification prediction. The loss value of the prediction result is then calculated and added to the loss values generated by Cls Head1 and Cls Head2. 

Notably, in this works, due to subtle differences in skin texture and color between monkeypox and other skin conditions, we set all activation functions to GELU. This is because GELU can capture small changes in the input and help alleviate the problem of gradient disappearance, which can provide more stable and accurate model training. Besides, we used a new regularization method——DropBlock[13]. DropBlock is a structured regularization method that randomly discards the entire continuous region on the feature map, which helps the network simulate information loss more effectively and enhances its robustness and generalization ability during training. Here, we used DropBlock for the output of stage 2 and stage 5 and set the size of the dropped area to 4x4 and 1x1 respectively.

\subsection{The principle of Attention-Based Local and Global Fusion Module}
The key to achieving high-precision semantic information extraction is to integrate multi-level features. In this work, we enriched and fused shallow and deep feature information using Local Context and Global Context. As shown in Figure 3c, ABLGFM uses attention modules to generate local context feature information l and global context feature information g, and fuses the input features (Flow and Fhigh) with l and g through Sigmod, element-wise product and element-wise add. l and g respectively correspond to Flow and Fhigh. Here, Flow is the output of the shallow network(Stage 2) and Fhigh is the output of the deeper network (Stage 5).

During the feature information propagation process, ABLGFM utilizes PWConv and Average pooling to down-sample Flow to the same image size as Fhigh. Then, the preliminary fusion of Flow and Fhigh is performed through element-wise add operation, and $Merge \in {R^{c \times h \times w}}$ is obtained. Next, we used a dual-branch attention modules to extract local context feature information $l \in {R^{c \times h \times w}}$ and global context feature information $g \in {R^{c \times 1 \times 1}}$ from Merge. Immediately after,  element-wise add operation is applied to obtain Weight. Then, the sigmoid function is used to scale Weight to the range of 0-1, and Weight' is obtained by subtracting weight from 1. Finally, to further enhance feature information on the original features, ABLGFM performs element-wise product between Weight',Fhigh, Weight and Flow, resulting in Fhigh' and Flow'. Finally, ABLGFM performs Add operation on Fhigh' and Flow' to obtain the final feature fusion result out. 

\textbf{Global Context Attention Module.} The key concept of the Global Context Attention Module is to generate weights by leveraging the relationship between each channel in the input features, which represents the importance of each channel. In this module, the adaptive average pooling operation is utilized to squeeze the spatial dimension of the input features, generating a feature with a size of ${R^{c \times 1 \times 1}}$. Subsequently, the generated feature undergoes PWConv, Batch Normalization (BN), and GELU operations to obtain the weights $g \in {R^{c \times 1 \times 1}}$. 

\textbf{Local Context Attention Module.} The key concept of the local context attention module is to extract fine-grained features from the preliminary fused feature map. As the global context attention module tends to focus on the dominant features in the entire feature map, disregarding a significant amount of fine-grained features. Therefore, to address this issue and extract fine-grained features, we incorporated the local context attention module in addition to the global context attention module. The local context attention module performs PWConv convolution, Batch Normalization (BN), and GELU operations on the preliminary fused feature map, resulting in the local contextual features, denoted as $l \in {R^{c \times h \times w}}$.

The process of ABLGFM can be summarized as the following formula:
\[ABLGFM\left\{ \begin{array}{l}
Merge = Fhigh \oplus (AveragePool(GELU(BN(PWconv(Flow)))))\\
g = BN(PWconv(GELU(BN(PWconv(AdaptiveAveragePool(Merge))))))\\
l = BN(PWconv(GELU(BN(PWconv(Merge)))))\\
weight' = Sigmod(l \oplus g)\\
Flow' = (AveragePool(GELU(BN(PWconv(Flow))))) \otimes (1 - weight')\\
Fhigh' = DropBlock(Fhigh) \otimes weight'\\
Out = Flow' + Fhigh'
\end{array} \right.\]

\subsection{Multiple auxiliary losses enhancement strategy}
Here, to further improve the diagnostic performance of the model without increasing the inference complexity, we propose a multiple auxiliary loss training strategy with a series of auxiliary classification heads (Figure 3b). This strategy can provide the model with loss information from different levels during the training phase, so as to better optimize the parameters of the backbone of the model. Importantly, during the inference phase of the model, we choose to directly discard these auxiliary classification heads. Therefore, compared with the network model before using this strategy, Fast-MpoxNet does not increase any computation during the inference phase, and will not reduce the speed of the model's inference. In addition, these auxiliary classification heads are flexible and can be inserted into different positions of the backbone.

\subsection{Lightweight strategy}
Although above mentioned improvements will greatly improve the network's ability to diagnose diseases, we find these improvements obviously increase the model parameter size of the network in practical applications and reduce the network's inference speed.

In the Fast-MpoxNet, the number of out channels in the feature map output by the last stage is 1024. Having more channels is beneficial for the final prediction results of the network, but it inevitably increases the model's parameters and decreases its inference speed. To alleviate the problem of decreased inference speed caused by ABLGFM, GELU and DropBlock, we made a trade-off and adjusted the number of out channels to 256.

\section{Results}
\begin{figure}[h]
    \centering
    \includegraphics[width=1\textwidth]{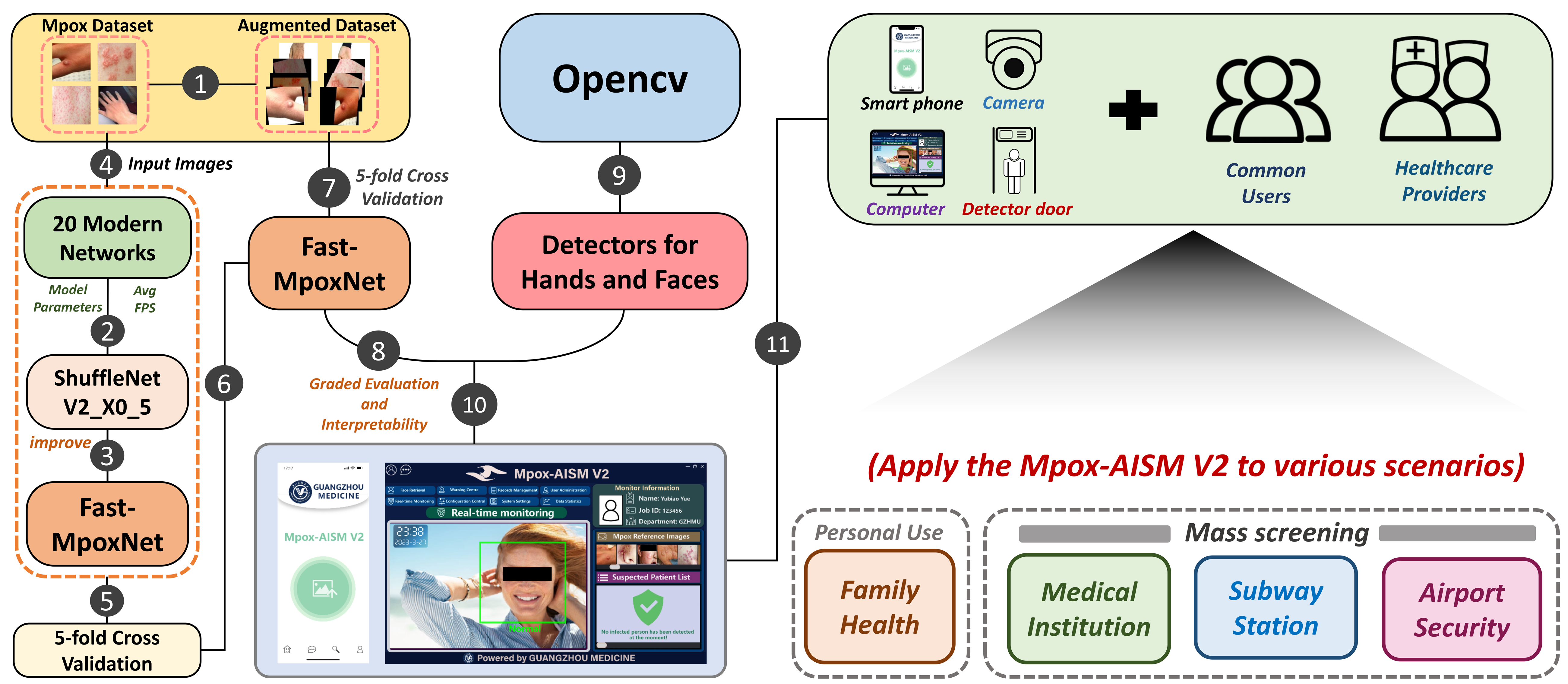}
    \caption{The workflow of our study.}
    \label{fig4}
\end{figure}
The initial step in our work was data collection, therefore we built the Mpox Dataset based on MSID and MSLD. Subsequently, data augmentation on the Mpox Dataset from the perspective of image clarity, lighting, contrast, and other aspects is performed. Following this, we selected 20 well-known networks, namely ShuffleNetV2\_X0\_5 [11], SqueezeNet1\_0 \cite{37}, Mobilevit\_xxs, Mobilevit\_xs \cite{38}, MNASNet0\_5 \cite{39}, EfficientNetB0, EfficientNetB1, EfficientNetB2 \cite{40}, MobileNet\_V2 \cite{41}, MobileNet\_V3\_small \cite{42}, GoogLeNet \cite{43}, ResNet18, ResNet34, ResNet50 \cite{44}, DesneNet121, DesneNet169, DesneNet201 \cite{45}, RegNet\_X\_400MF, RegNet\_Y\_400MF \cite{46}, Xception \cite{47}, performed a preliminary evaluation for parameter size and average FPS, and chose ShuffleNetV2\_X0\_5 as the baseline in this work. From there, we made reasonable improvements to ensure that the improved network performed well in terms of diagnosis performance, average FPS, and model parameters size. Here, we named the improved network Fast-MpoxNet and performed five-fold cross-validation on it, ShuffleNetV2\_X0\_5, and all other networks using the Mpox Dataset. After Fast-MpoxNet was evaluated, we also performed five-fold cross-validation on it using the augmented dataset in order to create useful applications. In addition, we also conducted a graded evaluation and model interpretability analysis of Fast-MpoxNet to ensure its suitability for real-world settings.

Our subsequent endeavor involved the development of specialized detectors for facial and hand regions by leveraging the OpenCV. Integrating Fast-MpoxNet with these detectors, we rolled out PC Mpox-AISM V2, an offline PC application. Concurrently, we introduced its mobile counterpart, Mobile Mpox-AISM V2, by directly integrating Fast-MpoxNet onto mobile platforms. The PC Mpox-AISM V2 is tailored for medical professionals. It adeptly and expeditiously identifies and diagnoses monkeypox lesions on facial and hand regions through real-time video. Its versatile design ensures seamless deployment across a range of devices, from personal computers to micro computing devices equipped with CPUs and cameras, and even specialized electronic detection doors. This adaptability positions PC Mpox-AISM V2 as an invaluable tool for medical staff, enabling them to conduct comprehensive screenings in high-risk environments during outbreaks, such as airports, subway stations, and hospitals and then discovered potential ealry-stage monkeypox patients. PC Mpox-AISM V2 can markedly improve the detection rate of early-stage monkeypox cases. Conversely, Mobile Mpox-AISM V2 is designed with the general populace in mind. Users can simply capture an image of the suspected lesion area and promptly receive a diagnostic assessment. The efficiency and compactness of Fast-MpoxNet ensure that Mpox-AISM V2 operates seamlessly even in the absence of GPU support, making it compatible with devices equipped with even modest CPUs. Figure 4 provides a comprehensive visual representation of our research methodology. The ensuing subsections will delve deeper into the specific experimental protocols and their corresponding results.

\subsection{Experimental setting}
Throughout our entire experiment, the epoch, batch size, learning rate, optimizer, and loss function were fixed. Table 4 provides a detailed summary of these hyper parameters, optimizer, and loss function. Additionally, we performed five-fold cross-validation for each model. Specifically, we divided the Mpox Dataset and the augmented Mpox Dataset (Aug dataset) into five equal parts respectively, selecting one part as the test set and the remaining four parts as the training set in turn. 
\begin{table}[]
\centering
\caption{Model Hyperparameters.}
\begin{tabular}{@{}ll@{}}
\toprule
\textbf{Hyperparameters} & \textbf{Values}    \\ \midrule
Optimizer                & Aadm               \\
Loss function            & Cross-entropy Loss \\
Batch size               & 32                 \\
Learning rate            & 0.0001             \\
Epochs                   & 100                \\ \bottomrule
\end{tabular}
\end{table}

\begin{table}[]
\centering
\caption{Mpox Dataset splitting in details.}
\begin{tabular}{@{}lllllllll@{}}
\toprule
               & \multicolumn{2}{l}{\textbf{Monkeypox}} & \multicolumn{2}{l}{\textbf{Chickenpox}} & \multicolumn{2}{l}{\textbf{Measles}} & \multicolumn{2}{l}{\textbf{Normal}} \\ \midrule
               & Train              & Test              & Train               & Test              & Train             & Test             & Train             & Test            \\
\textbf{Fold1} & 305                & 76                & 86                  & 21                & 73                & 18               & 235               & 58              \\
\textbf{Fold2} & 305                & 76                & 86                  & 21                & 73                & 18               & 235               & 58              \\
\textbf{Fold3} & 305                & 76                & 86                  & 21                & 73                & 18               & 235               & 58              \\
\textbf{Fold4} & 305                & 76                & 86                  & 21                & 73                & 18               & 235               & 58              \\
\textbf{Fold5} & 304                & 77                & 84                  & 23                & 72                & 19               & 232               & 61              \\ \bottomrule
\end{tabular}
\end{table}
The distribution of images in the five folds for both the Mpox Dataset and Aug Dataset is presented in Table 5 and Table 6, respectively. The experimental environment for this work was based on Ubuntu22.04, Nvidia 3090ti and Pytorch framework. In addition, when testing the average FPS metric of the model, we used a laptop computer with an Intel(R) Core(TM) i7-10870H CPU @2.20Hz.

\begin{table}[]
\centering
\caption{Augmented Mpox Dataset splitting in details.}
\begin{tabular}{@{}lllllllll@{}}
\toprule
               & \multicolumn{2}{l}{\textbf{Monkeypox}} & \multicolumn{2}{l}{\textbf{Chickenpox}} & \multicolumn{2}{l}{\textbf{Measles}} & \multicolumn{2}{l}{\textbf{Normal}} \\ \midrule
               & Train              & Test              & Train               & Test              & Train             & Test             & Train             & Test            \\
\textbf{Fold1} & 2134               & 533               & 1884                & 470               & 1893              & 473              & 1863              & 465             \\
\textbf{Fold2} & 2134               & 533               & 1884                & 470               & 1893              & 473              & 1863              & 465             \\
\textbf{Fold3} & 2134               & 533               & 1884                & 470               & 1893              & 473              & 1863              & 465             \\
\textbf{Fold4} & 2134               & 533               & 1884                & 470               & 1893              & 473              & 1863              & 465             \\
\textbf{Fold5} & 2132               & 535               & 1880                & 474               & 1892              & 474              & 1860              & 468             \\ \bottomrule
\end{tabular}
\end{table}

\subsection{Model comparison and ablation experiment}
To demonstrate that Fast-MpoxNet has better diagnostic capabilities and practicality than the original ShuffleNetV2, we conducted comprehensive evaluations of ShufflenetV2 and Fast-MpoxNet with and without ImageNet pre-trained weights using Mpox Dataset and seven evaluation metrics under consistent experimental conditions. The models under these four conditions are referred to as ShuffleNetV2, ShuffleNetV2 (ImageNet), Fast-MpoxNet and Fast-MpoxNet (ImageNet), respectively. Additionally, we plotted confusion matrices for Fast-MpoxNet and Fast-MpoxNet (ImageNet).

According to application requirements, we first compared the average FPS and model parameters size between ShuffleNetV2 and Fast-MpoxNet. Using Equation(6) with N set to 100 and computed on an Intel(R) Core(TM) i7-10870H CPU @2.20Hz, Fast-MpoxNet and ShuffleNetV2 achieved average FPS of 68 and 74, respectively. Using Pytorch Framework, Fast-MpoxNet and ShuffleNetV2 had model parameter quantities of 0.27M and 0.35M, respectively. Subsequently, we conducted five-fold cross-validation on Fast-MpoxNet and ShuffleNetV2 using the Mpox Dataset. Figure 5a shows the accuracy of ShuffleNet, ShuffleNetV2 (ImageNet), Fast-MpoxNet, and Fast-MpoxNet (ImageNet) on the test set of the five folds. The accuracy of Fast-MpoxNet on the test set of the five folds was 80.35\%, 77.46\%, 78.61\%, 78.61\%, and 82.78\%, respectively. The accuracy of ShuffleNetV2 on the test set of the five folds was 76.30\%, 74.57\%, 76.88\%, 76.88\%, and 75.56\%, respectively. The accuracy of Fast-MpoxNet (ImageNet) on the test set of the five folds was 97.11\%, 92.49\%, 95.95\%, 90.75\%, and 95.00\%, respectively. The accuracy of ShuffleNetV2 (ImageNet) on the test set of the five folds was 92.49\%, 90.75\%, 95.38\%, 89.60\% and 93.33\%  respectively. In Figure 5a, it can be visually observed that Fast-MpoxNet outperforms ShuffleNetV2 in terms of accuracy in any case. Similarly, we conducted ablation experiments under the same experimental conditions. The average accuracy improvement results brought about by different improvements are plotted in Figure 5b. Among these improvements, the improvement brought by ABLGFM is the most obvious. It increases the accuracy of the network from 76.04\% to 78.17\%. Overall, with these improvements, Fast-MpoxNet can finally be Get 79.56\% accuracy. In the case of using pre-trained weights, the accuracy of Fast-MpoxNet can reach 94.26\%. Compared with ShuffleNetV2, it has increased by 3.52\% and 1.95\% respectively.
\begin{figure}[h]
    \centering
    \includegraphics[width=1\textwidth]{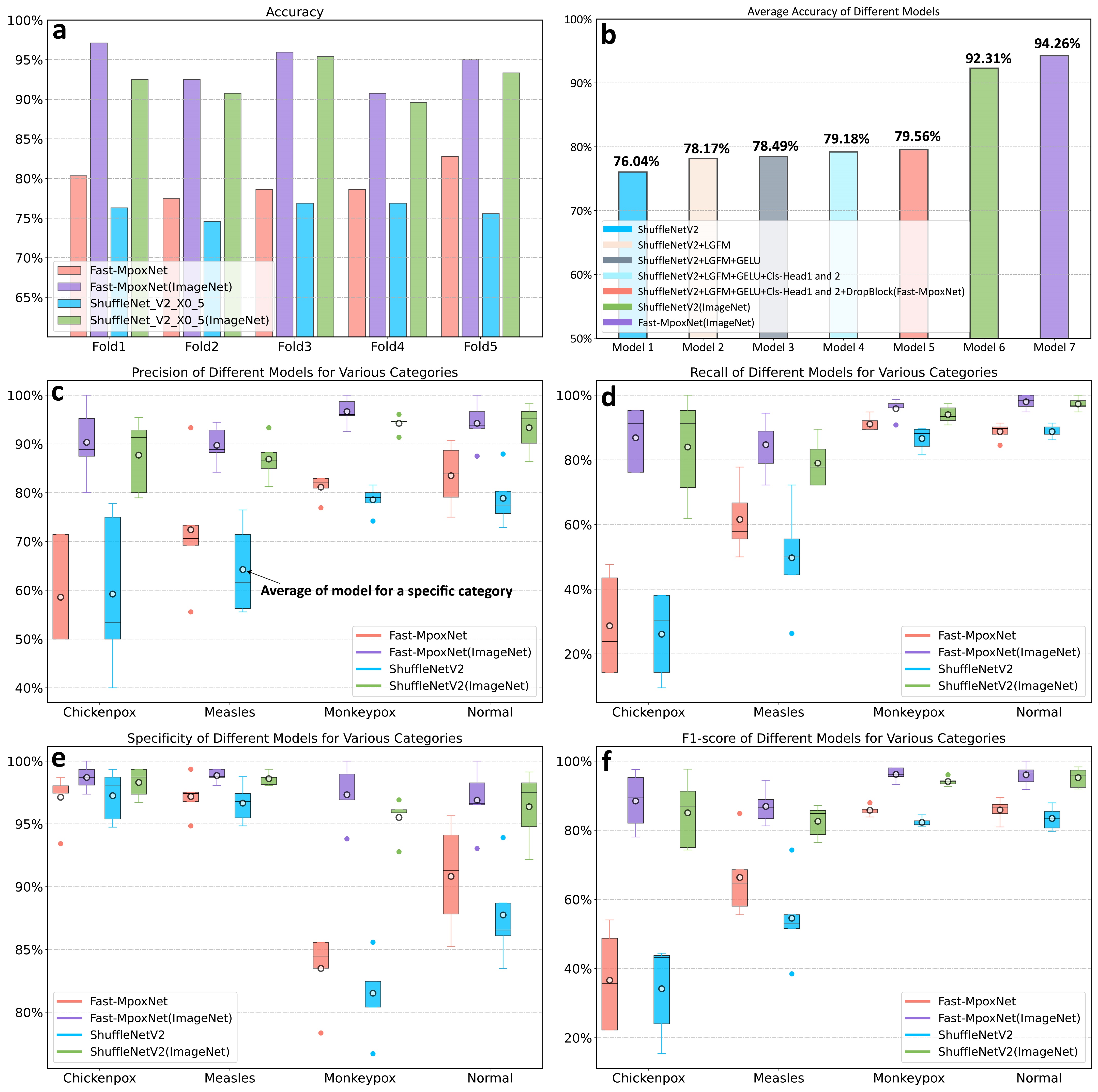}
    \caption{a: The Accuracy between four models on each fold. b: The average Accuracy of different models in the ablation experiment. c: The Precision between four models for various diseases. d: The Recall between four models for various diseases. e: The Specificity between four models for various diseases. f: The F1-score between four models for various diseases.}
    \label{fig5}
\end{figure}
According to our evaluation strategy, we, in the box plot (Figure 5c-f), plotted the Precision, Recall, Specificity and F1-score between the four models for each category. In these figures, we also plotted their averaged results for each class in each fold (hollow scatter points in the figure). It can be intuitively seen from the height of each box that Fast-MpoxNet exhibits a more stable state. From the average point of view, Fast-MpoxNet performs better for each category of index results in any case. Looking at the results of each indicator comprehensively, Fast-MpoxNet has a clear advantage in distinguishing monkeypox from other images.

We also plotted confusion matrices of different models to show the performance of Fast-MpoxNet under various situations in Figure 6. In these matrices, the proportion of values on the diagonal to the total value of their respective columns represents the Recall. A higher Recall for a particular class indicates that the model can better differentiate that class from other classes, and that the samples of that class are less prone to misclassification or missed detection. When ImageNet weights are not used, the highest and lowest Recall of monkeypox for Fast-MpoxNet achieved 94.81\% (Figure 6a) and 89.47\% (Figure 6a) respectively. When ImageNet weights are used, the highest and lowest Recall of monkeypox for Fast-MpoxNet (ImageNet) achieved 97.37\% (Figure 6b) and 90.79\%  (Figure 6b) respectively. In addition, the best Fast-Monkeypox(ImageNet) also exhibits excellent Recall performance in all three other categories. Compared with ShuffleNetV2(Figure 6c-d), Fast-MpoxNet is still better in these confusion matrices.
\begin{figure}[h]
    \centering
    \includegraphics[width=1\textwidth]{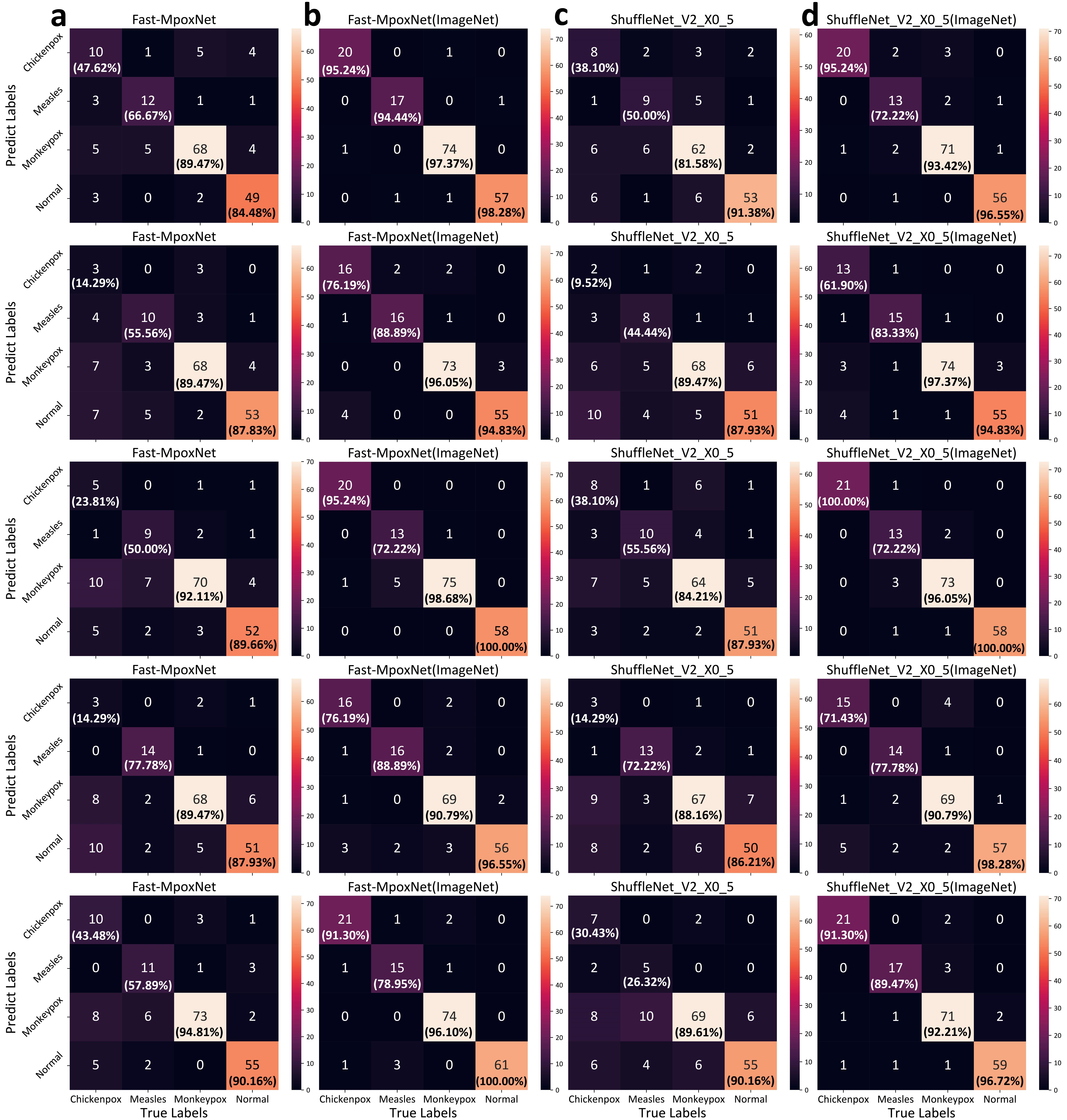}
    \caption{The confusion matrices of the four models. From top to bottom, these matrices correspond to Fold1 to Fold5, respectively. a: The confusion matrices of Fast-MpoxNet. b: The confusion matrices of Fast-MpoxNet(ImageNet). c: The confusion matrices of ShuffleNet. d: The confusion matrices of ShuffleNet(ImageNet).}
    \label{fig6}
\end{figure}
Finally, we summarized the average Precision, average Recall, average Specificity and average F1-score of Fast-MpoxNet and ShuffleNetV2 in Table 7. It is evident that despite a lightly decrease in average FPS, Fast-MpoxNet greatly outperformed ShuffleNetV2 in term of diagnosis performance. As mentioned earlier, a model applied to various real-world settings should not just be able to lead on a single metric, but to achieve a good trade-off between diagnostic ability, average FPS, and model parameter size. From all these results, our proposed Fast-MpoxNet is more suitable for practical applications.

\begin{table}[]
\centering
\caption{Comparison of different metrics between ShuffleNetV2, ShuffleNetV2 (ImageNet), Fast-MpoxNet, and Fast-MpoxNet (ImageNet).}
\begin{tabular}{@{}lllll@{}}
\toprule
\textbf{Model}                     & \textbf{Average   Precision} & \textbf{Average   Recall} & \textbf{Average   Specificity} & \textbf{Average   F1-score} \\ \midrule
ShuffleNetV2                       & 70.22\%(±4.81)               & 62.78\%(±3.56)            & 90.79\%(±0.54)                 & 63.61\%(±3.44)              \\
\textbf{Fast-MpoxNet}              & \textbf{73.90\%(±3.92)}      & \textbf{67.52\%(±3.84)}   & \textbf{92.16\%(±0.81)}        & \textbf{67.65\%(±3.48)}     \\
ShuffleNetV2   (ImageNet)          & 89.90\%(±2.88)               & 88.56\%(±3.51)            & 97.19\%(±0.79)                 & 89.23\%(±2.69)              \\
\textbf{Fast-MpoxNet   (ImageNet)} & \textbf{92.73\%(±3.13)}      & \textbf{91.31\%(±2.86)}   & \textbf{97.94\%(±0.81)}        & \textbf{91.88\%(±2.88)}     \\ \bottomrule
\end{tabular}
\end{table}

\subsection{Comparison with other popular networks}
\begin{table}[]
\caption{Comparison of seven metrics between Fast-MpoxNet (ImageNet) and other popular networks.}
\resizebox{1\textwidth}{!}{
\begin{tabular}{@{}lllllllll@{}}
\toprule
\textbf{\begin{tabular}[c]{@{}l@{}}Model   \\ Name\end{tabular}} & \textbf{\begin{tabular}[c]{@{}l@{}}Practicality\\  score\end{tabular}} & \textbf{Avg   FPS} & \textbf{\begin{tabular}[c]{@{}l@{}}Model\\  Params\end{tabular}} & \textbf{\begin{tabular}[c]{@{}l@{}}Mean\\Accuracy\end{tabular}} & \textbf{\begin{tabular}[c]{@{}l@{}}Precision\\ (Mpox)\end{tabular}} & \textbf{\begin{tabular}[c]{@{}l@{}}Recall\\(Mpox)\end{tabular}} & \textbf{\begin{tabular}[c]{@{}l@{}}Specificity\\ (Mpox)\end{tabular}} & \textbf{\begin{tabular}[c]{@{}l@{}}F1-score\\  (Mpox)\end{tabular}} \\ \midrule
Fast-MpoxNet                                                           & 0.8291                                                                      & 68                 & 0.27M                                                                 & 94.26\%   (±2.32)                                                      & \begin{tabular}[c]{@{}l@{}}96.63\%(±2.56)\end{tabular}           & 95.80\%   (±2.68)                                                      & \begin{tabular}[c]{@{}l@{}}97.32\%(±2.12)\end{tabular}             & 96.17\%   (±1.78)                                                        \\
SqueezeNet1\_0                                                         & 0.5429                                                                      & 40                 & 0.74M                                                                 & 93.21\%   (±1.74)                                                      & \begin{tabular}[c]{@{}l@{}}95.17\%(±3.16)\end{tabular}           & 95.00\%   (±2.55)                                                      & \begin{tabular}[c]{@{}l@{}}96.15\%   (±2.58)\end{tabular}          & 95.02\%   (±1.38)                                                        \\
Mobilevit\_xxs                                                         & 0.5847                                                                      & 25                 & 1.3M                                                                  & 95.06\%   (±0.91)                                                      & \begin{tabular}[c]{@{}l@{}}96.66\%    (±2.27)\end{tabular}           & 96.32\%   (±1.54)                                                      & \begin{tabular}[c]{@{}l@{}}97.36\%  (±1.91)\end{tabular}          & 96.46\%   (±0.95)                                                        \\
Mobilevit\_xs                                                          & 0.5949                                                                      & 13                 & 2.3M                                                                  & 96.20\%   (±1.15)                                                      & \begin{tabular}[c]{@{}l@{}}97.67\%    (±1.68)\end{tabular}           & 97.11\%   (±2.11)                                                      & \begin{tabular}[c]{@{}l@{}}98.16\%     (±1.37)\end{tabular}          & 97.36\%   (±0.85)                                                        \\
MobileNet\_V3\_small                                                   & 0.7843                                                                      & 60                 & 1.52M                                                                 & 94.26\%   (±1.08)                                                      & \begin{tabular}[c]{@{}l@{}}96.61\%    (±1.75)\end{tabular}           & 96.07\%   (±1.43)                                                      & \begin{tabular}[c]{@{}l@{}}97.32\%  (±1.40)\end{tabular}          & 96.32\%   (±0.68)                                                        \\
MobileNet\_V2                                                          & 0.6159                                                                      & 22                 & 2.23M                                                                 & 95.25\%   (±0.48)                                                      & \begin{tabular}[c]{@{}l@{}}97.89\%    (±1.56)\end{tabular}           & 96.06\%   (±0.83)                                                      & \begin{tabular}[c]{@{}l@{}}98.36\%    (±1.24)\end{tabular}          & 96.96\%   (±0.51)                                                        \\
MNASNet0\_5                                                            & 0.3961                                                                      & 54                 & 0.94M                                                                 & 89.42\%   (±1.92)                                                      & \begin{tabular}[c]{@{}l@{}}96.80\%    (±1.24)\end{tabular}           & 85.55\%   (±4.19)                                                      & \begin{tabular}[c]{@{}l@{}}97.78\% (±0.92)\end{tabular}          & 90.75\%   (±0.02)                                                        \\
EfficientNetB0                                                         & 0.5578                                                                      & 16                 & 4.01M                                                                 & 95.63\%   (±1.10)                                                      & \begin{tabular}[c]{@{}l@{}}97.34\%    (±1.47)\end{tabular}           & 95.80\%   (±1.54)                                                      & \begin{tabular}[c]{@{}l@{}}97.96\% (±1.13)\end{tabular}          & 96.56\% (±1.42)                                                        \\
EfficientNetB1                                                         & 0.6231                                                                      & 12                 & 6.52M                                                                 & 96.10\%   (±0.69)                                                      & \begin{tabular}[c]{@{}l@{}}98.66\%    (±0.02)\end{tabular}           & 96.85\%   (±1.33)                                                      & \begin{tabular}[c]{@{}l@{}}98.98\%  (±0.02)\end{tabular}          & 97.75\%   (±0.69)                                                        \\
EfficientNetB2                                                         & 0.5884                                                                      & 11                 & 7.71M                                                                 & 96.21\%   (±1.50)                                                      & \begin{tabular}[c]{@{}l@{}}98.40\%     (±1.00)\end{tabular}           & 96.06\%   (±1.66)                                                      & \begin{tabular}[c]{@{}l@{}}98.78\% (±0.77)\end{tabular}          & 97.21\%   (±1.00)                                                        \\
RegNet\_X\_400MF                                                       & 0.6804                                                                      & 38                 & 5.10M                                                                 & 95.06\%   (±0.90)                                                      & \begin{tabular}[c]{@{}l@{}}96.66\%     (±2.28)\end{tabular}           & 96.85\%   (±1.33)                                                      & \begin{tabular}[c]{@{}l@{}}97.33\% (±1.92)\end{tabular}          & 96.73\%   (±0.97)                                                        \\
RegNet\_Y\_400MF                                                       & 0.7203                                                                      & 35                 & 3.90M                                                                 & 95.52\%   (±1.19)                                                      & \begin{tabular}[c]{@{}l@{}}97.89\%     (±1.00)\end{tabular}           & 96.59\%   (±1.34)                                                      & \begin{tabular}[c]{@{}l@{}}98.38\% (±0.81)\end{tabular}          & 97.23\%   (±0.50)                                                        \\
ResNet18                                                               & 0.5285                                                                      & 28                 & 11.18M                                                                & 94.36\%   (±1.34)                                                      & \begin{tabular}[c]{@{}l@{}}95.81\%    (±1.81)\end{tabular}           & 95.27\%   (±1.35)                                                      & \begin{tabular}[c]{@{}l@{}}96.73\%(±1.54)\end{tabular}            & 95.53\%   (±1.12)                                                        \\
ResNet34                                                               & 0.5073                                                                      & 20                 & 21.29M                                                                & 94.26\%   (±1.06)                                                      & \begin{tabular}[c]{@{}l@{}}96.62\%    (±2.30)\end{tabular}           & 95.28\%   (±2.44)                                                      & \begin{tabular}[c]{@{}l@{}}97.34\% (±1.92)\end{tabular}            & 95.90\%   (±1.42)                                                        \\
ResNet50                                                               & 0.5424                                                                      & 10                 & 23.52M                                                                & 95.75\%   (±0.73)                                                      & \begin{tabular}[c]{@{}l@{}}97.93\%     (±2.08)\end{tabular}           & 95.80\%   (±2.25)                                                      & \begin{tabular}[c]{@{}l@{}}98.35\% (±1.67)\end{tabular}          & 96.81\%   (±0.67)                                                        \\
DenseNet121                                                            & 0.4733                                                                      & 11                 & 6.96M                                                                 & 94.94\%   (±1.33)                                                      & \begin{tabular}[c]{@{}l@{}}95.65\%     (±1.98)\end{tabular}           & 97.37\%   (±2.04)                                                      & \begin{tabular}[c]{@{}l@{}}96.51\%(±1.69)\end{tabular}            & 96.48\%   (±1.42)                                                        \\
DenseNet169                                                            & 0.4695                                                                      & 9                  & 12.49M                                                                & 95.29\%   (±1.03)                                                      & \begin{tabular}[c]{@{}l@{}}96.85\%    (±1.77)\end{tabular}           & 96.59\%   (±1.34)                                                      & \begin{tabular}[c]{@{}l@{}}96.91\% (±1.72)\end{tabular}          & 96.34\%   (±0.97)                                                        \\
DenseNet201                                                            & 0.5052                                                                      & 6                  & 18.10M                                                                & 95.51\%   (±1.37)                                                      & \begin{tabular}[c]{@{}l@{}}97.87\%    (±1.02)\end{tabular}           & 95.54\%   (±1.79)                                                      & \begin{tabular}[c]{@{}l@{}}98.36\% (±0.83)\end{tabular}          & 96.67\%   (±0.77)                                                        \\
GoogLeNet                                                              & 0.5504                                                                      & 22                 & 9.94M                                                                 & 94.83\%   (±1.45)                                                      & \begin{tabular}[c]{@{}l@{}}96.14\%    (±2.24)\end{tabular}           & 96.85\%   (±1.35)                                                      & \begin{tabular}[c]{@{}l@{}}96.92\%(±1.86)\end{tabular}            & 96.47\%   (±1.33)                                                        \\
Xception                                                               & 0.6473                                                                      & 23                 & 20.76M                                                                & 95.97\%   (±1.12)                                                      & \begin{tabular}[c]{@{}l@{}}98.38\%   (±0.54)\end{tabular}           & 95.54\%   (±0.65)                                                      & \begin{tabular}[c]{@{}l@{}}98.78\% (±0.42)\end{tabular}          & 96.94\%   (±0.54)                                                        \\ \bottomrule
\end{tabular}}
\end{table}
In this subsection, we performed five-fold cross-validation on Fast-MpoxNet and a number of previously related networks as well as popular lightweight networks using the Mpox Dataset. Here, all networks had loaded pre-trained weights from ImageNet. As shown in the Table 8, we evaluate models using seven metrics. Here, we focused on the performance of networks for the monkeypox images. Fast-MpoxNet achieved Precision of 96.63\% (±2.56), Recall of 95.80\% (±2.68), Specificity of 97.32\% (±2.12), and F1-score of 96.17\% (±1.78). These metrics demonstrate that Fast-MpoxNet still has excellent monkeypox diagnostic ability under the conditions of 0.27M parameters and 68 average FPS. Moreover, the parameter size of Fast-MpoxNet is only 17.7\% of famous MobileNet\_V3\_small, but it is on par with it in terms of average accuracy and has increased in average FPS by 13.33\%. Theoretically, in a large-scale screening environment, Fast-MpoxNet can diagnose eight more individuals per second than MobileNet\_V3\_small. Due to its crucial role in settings requiring large-scale screening of populations, FPS has been the focus of our attention, and we have therefore generated a comparative diagram specifically dedicated to this metric (Figure 7). In Figure 7a, the closer the network is to the upper right of the scatter plot, the better it is. Clearly, Fast-MpoxNet is closer to the upper right of the scatter plot, which means that Fast-MpoxNet achieves a good balance between Avg FPS and Mean Accuracy. In Figure 7b, the closer the network is to the upper left of the scatter plot, the better it is. Clearly, Fast-MpoxNet is closer to the upper left of the scatter plot, which means that Fast-MpoxNet achieves a good balance between Avg FPS and Model Parameters. Besides, according Equation (11), We calculated the practicality scores of the models. The computational results indicated that Fast-MpoxNet obtained the highest score of 0.8291, implying its superior practicality among the tested models. Therefore, based on the above results, it can be concluded that Fast-MpoxNet is more suitable for diagnosing monkeypox patients in various practical settings.
\begin{figure}[h]
    \centering
    \includegraphics[width=1\textwidth]{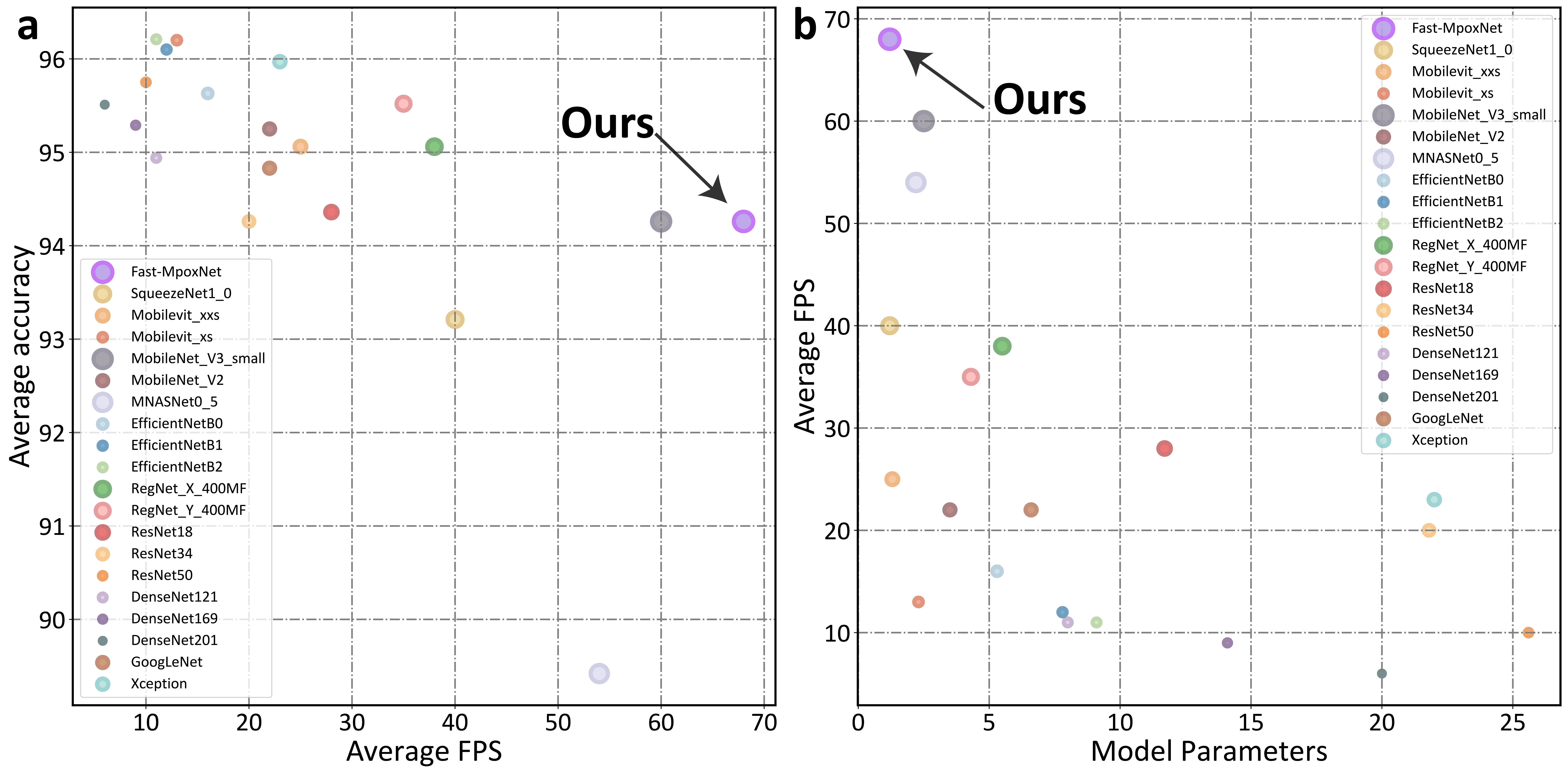}
    \caption{a: The relationship between average Accuracy and average FPS. b: The relationship between model parameters size and average FPS.}
    \label{fig7}
\end{figure}
After conducting comprehensive five-fold cross-validation via Mpox Dataset for these 20 popular networks, to further enhance the robustness of the Fast-MpoxNet, we subsequently performed five-fold cross-validation on Fast-MpoxNet using the Aug Dataset according to practical deployment requirements. As before, we first load the pre-trained weights on ImageNet for each model. In addition, we also repeated the five-fold cross-validation on ResNet34, MobileNet\_V3\_small, Densenet201, MNASNet0\_5 and SqueezeNet1\_0 using the Aug Dataset to highlight the advantage of Fast-MpoxNet on a larger dataset. The experimental results were summarized in the Table 9. As shown in the Table 9, when tested on a larger dataset, Fast-MpoxNet only slightly lagged behind the other three models (MobileNet\_V3\_small, Densenet201, ResNet34) on the five metrics, yet it exhibited an average FPS of 68 and a parameter size of only 0.27M. Most importantly, the practicality score of Fast-MpoxNet outperformed the other five models with a score of 0.7976 which demonstrated the advantage of Fast-MpoxNet in practical deployment. Furthermore, the performance of Fast-MpoxNet on the Aug Dataset also indicates to some extent that if a truly large-scale Mpox Dataset is available in the future, Fast-MpoxNet can be trained to achieve better diagnostic performance. We denoted Fast-MpoxNet using Aug Dataset and pre-trained weights as Fast-MpoxNet (Aug and ImageNet).
\begin{table}[]
\centering
\caption{Comparison of Accuracy, Precision, Recall, Specificity, F1-score, Practicality Score on Augmented Mpox Dataset.}
\begin{tabular}{@{}lllllllll@{}}
\toprule
\textbf{Model   Name}    & \textbf{Metrics} & \textbf{Fold1} & \textbf{Fold2} & \textbf{Fold3} & \textbf{Fold4} & \textbf{Fold5} & \textbf{Average} & \textbf{Score} \\ \midrule
Fast-MpoxNet(68)         & Accuracy         & 98.51\%        & 98.40\%        & 98.04\%        & 98.66\%        & 98.41\%        & 98.40\%(±0.20)   & 0.7976         \\
                         & Precision        & 98.52\%        & 98.41\%        & 98.05\%        & 98.67\%        & 98.40\%        & 98.41\%(±0.20)   &                \\
                         & Recall           & 98.53\%        & 98.48\%        & 98.07\%        & 98.68\%        & 98.44\%        & 98.44\%(±0.20)   &                \\
                         & Specificity      & 99.50\%        & 99.47\%        & 99.35\%        & 99.55\%        & 99.48\%        & 99.47\%(±0.07)   &                \\
                         & F1-score         & 98.52\%        & 98.43\%        & 98.05\%        & 98.67\%        & 98.41\%        & 98.42\%(±0.20)   &                \\
ResNet34(20)             & Accuracy         & 98.71\%        & 98.40\%        & 98.40\%        & 98.56\%        & 98.31\%        & 98.48\%(±0.09)   & 0.5147         \\
                         & Precision        & 98.75\%        & 98.42\%        & 98.45\%        & 98.55\%        & 98.32\%        & 98.50\%(±0.15)   &                \\
                         & Recall           & 98.71\%        & 98.44\%        & 98.40\%        & 98.61\%        & 98.34\%        & 98.50\%(±0.14)   &                \\
                         & Specificity      & 99.57\%        & 99.47\%        & 99.47\%        & 99.52\%        & 99.44\%        & 99.49\%(±0.05)   &                \\
                         & F1-score         & 98.73\%        & 98.42\%        & 98.42\%        & 98.58\%        & 98.33\%        & 98.50\%(±0.14)   &                \\
MobileNet\_V3\_small(60) & Accuracy         & 98.56\%        & 98.40\%        & 98.35\%        & 98.56\%        & 98.51\%        & 98.48\%(±0.14)   & 0.777          \\
                         & Precision        & 98.56\%        & 98.41\%        & 98.35\%        & 98.57\%        & 98.53\%        & 98.48\%(±0.09)   &                \\
                         & Recall           & 98.58\%        & 98.47\%        & 98.41\%        & 98.60\%        & 98.56\%        & 98.52\%(±0.07)   &                \\
                         & Specificity      & 99.52\%        & 99.47\%        & 99.45\%        & 99.52\%        & 99.50\%        & 99.49\%(±0.03)   &                \\
                         & F1-score         & 98.57\%        & 98.43\%        & 98.37\%        & 98.58\%        & 98.54\%        & 98.50\%(±0.08)   &                \\
DenseNet201(6)           & Accuracy         & 98.76\%        & 98.92\%        & 99.02\%        & 98.97\%        & 98.82\%        & 98.90\%(±0.10)   & 0.6000         \\
                         & Precision        & 98.76\%        & 98.92\%        & 99.02\%        & 98.97\%        & 98.82\%        & 98.90\%(±0.10)   &                \\
                         & Recall           & 98.79\%        & 98.96\%        & 99.03\%        & 99.00\%        & 98.85\%        & 98.93\%(±0.09)   &                \\
                         & Specificity      & 99.59\%        & 99.64\%        & 99.68\%        & 99.65\%        & 99.61\%        & 99.63\%(±0.03)   &                \\
                         & F1-score         & 98.77\%        & 98.94\%        & 99.02\%        & 98.98\%        & 98.83\%        & 98.91\%(±0.09)   &                \\
MNASNet0\_5(54)          & Accuracy         & 98.30\%        & 98.45\%        & 97.94\%        & 98.15\%        & 98.15\%        & 98.20\%(±0.17)   & 0.6236         \\
                         & Precision        & 98.30\%        & 98.47\%        & 97.98\%        & 98.17\%        & 98.17\%        & 98.22\%(±0.16)   &                \\
                         & Recall           & 98.34\%        & 98.50\%        & 97.97\%        & 98.20\%        & 98.17\%        & 98.24\%(±0.18)   &                \\
                         & Specificity      & 99.43\%        & 99.48\%        & 99.31\%        & 99.38\%        & 99.38\%        & 99.40\%(±0.06)   &                \\
                         & F1-score         & 98.32\%        & 98.48\%        & 97.97\%        & 98.18\%        & 98.17\%        & 98.22\%(±0.17)   &                \\
SqueezeNet1\_0(40)       & Accuracy         & 98.76\%        & 98.92\%        & 99.02\%        & 98.97\%        & 98.82\%        & 97.44\%(±0.12)   & 0.2194         \\
                         & Precision        & 97.34   \%     & 97.33\%        & 97.52\%        & 97.60\%        & 97.56\%        & 97.47\%(±0.11)   &                \\
                         & Recall           & 97.38\%        & 97.29\%        & 97.51\%        & 97.65\%        & 97.62\%        & 97.49\%(±0.14)   &                \\
                         & Specificity      & 99.10\%        & 99.09\%        & 99.16\%        & 99.19\%        & 99.18\%        & 99.14\%(±0.04)   &                \\
                         & F1-score         & 97.36\%        & 97.31\%        & 97.50\%        & 97.62\%        & 97.58\%        & 97.47\%(±0.12)   &                \\ \bottomrule
\end{tabular}
\end{table}
We plotted the confusion matrix and receiver operating characteristic curve (ROC) in Figure 8 to further analyze the performance of Fast-MpoxNet (Aug and ImageNet) in diagnosing various diseases. Figure 8a and Figure 8c represent the confusion matrix and ROC when it achieves the best accuracy. The results show that the Recall of Fast-MpoxNet (Aug and ImageNet) for the four types of diseases can reach 98.29\%, 98.11\%, 98.12\% and 100\%, respectively. In the ROC curves (Figure 8c), the larger the area under the curve (AUC) for a certain class, the better the model can distinguish this class. The AUC for Monkeypox was about 0.9990, and the corresponding AUC values for the other three classes were also over 0.99, indicating that the Fast-MpoxNet (ImageNet and Aug) was able to effectively distinguish different classes. As this was a multi-classification task, both the macro-average ROC and micro-average ROC were plotted. The macro-average ROC is the ROC curve obtained by simply averaging the ROC curves for each class, which equally considers the contribution of each class in the overall evaluation. Its calculation method is to simply average the true positive rate and false positive rate for each class. The micro-average ROC is the ROC curve calculated by summing the true positive and false positive counts for all classes. It is suitable for situations where the sample sizes of different classes are imbalanced in multi-classification problems. The corresponding AUC values for both the macro-average and micro-average ROCs were over 99\%, indicating that the Fast-MpoxNet (ImageNet and Aug) achieved high predictive performance on multiple classes and maintained good balance between different classes.  Figure 8b represents the confusion matrix when it achieves the worst Accuracy. It shows that Fast-MpoxNet (Aug and ImageNet) can reach more than 97\% of Recall for the four types of diseases. Likewise, the ROC in Figure 8d proves that even in the worst case, Fast-MpoxNet (ImageNet and Aug) with an AUC above 0.99 for each category still maintains a good diagnostic ability for each category.
\begin{figure}[h]
    \centering
    \includegraphics[width=1\textwidth]{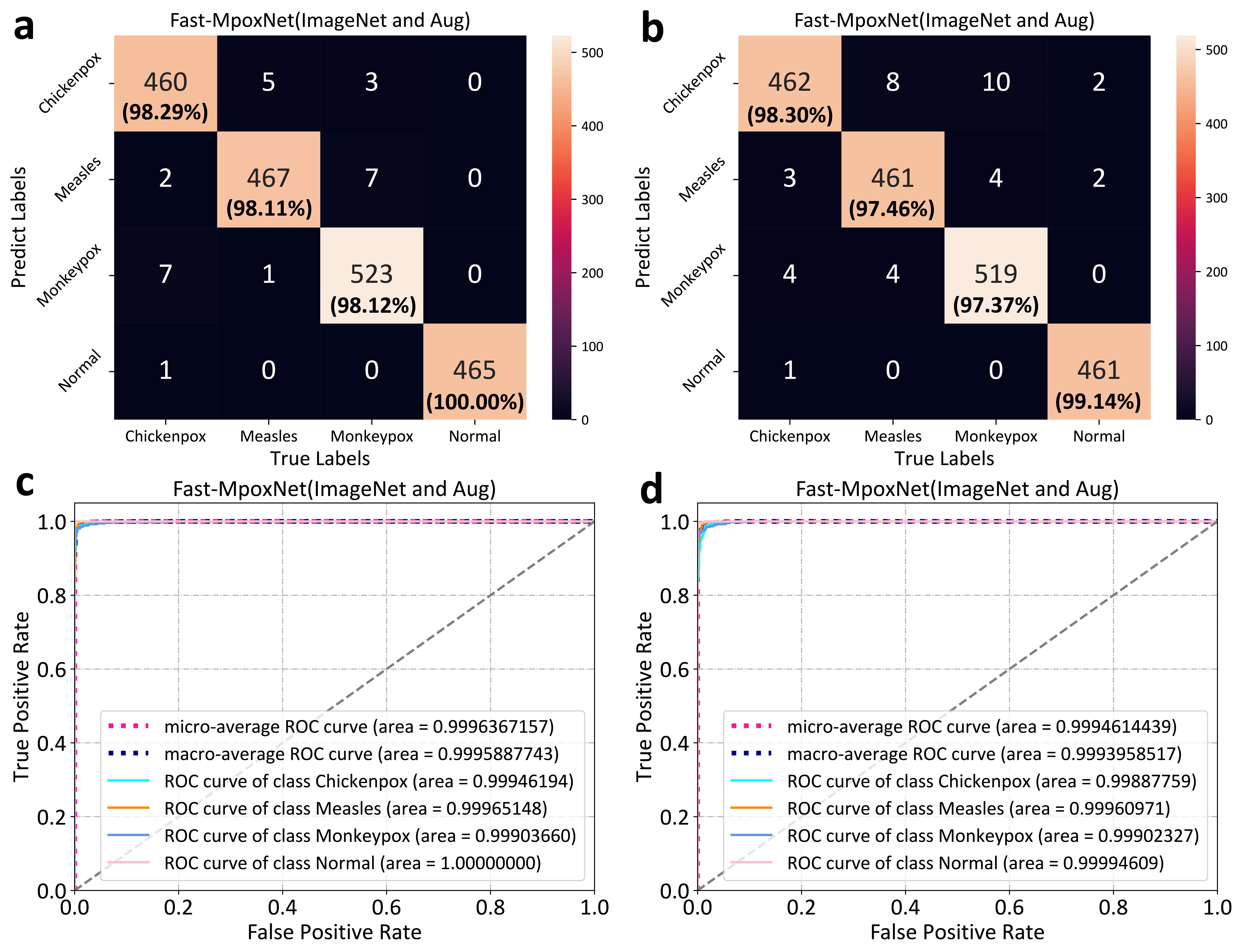}
    \caption{a: The confusion matrix of Fast-MpoxNet(ImageNet and Aug). b: The ROC curve of Fast-MpoxNet(ImageNet and Aug). c: The ROC curve of Fast-MpoxNet with 97.11\% Accuracy. d: The ROC curve of Fast-MpoxNet with 80.12\% Accuracy.}
    \label{fig8}
\end{figure}
\subsection{Graded evaluation of Fast-MpoxNet}
To apply the monkeypox diagnosis system to various real-world settings, it is necessary to conduct a graded evaluation of monkeypox. Graded evaluation means Fast-MpoxNet test monkeypox images from different stages and different human body parts separately. In practical applications, graded evaluation results can provide data support for users in various settings to assist them in further confirming the condition. This idea has already been adopted in our previous work and has also been considered by the Stanford medical team in [16]. Prior to graded evaluation, dermatologists first divided the monkeypox images in the test set of each fold into seven portions: arm, face, hand, leg, trunk, feet, and close-range (i.e., cannot determine specific parts). In addition, experts also divided the images into two stages based on dataset characteristics and the clinical development trend of monkeypox: Early-stage and Later-stage. The samples of seven portions and two stages from Monkeypox images are shown in Figure 9. 
\begin{figure}[h]
    \centering
    \includegraphics[width=1\textwidth]{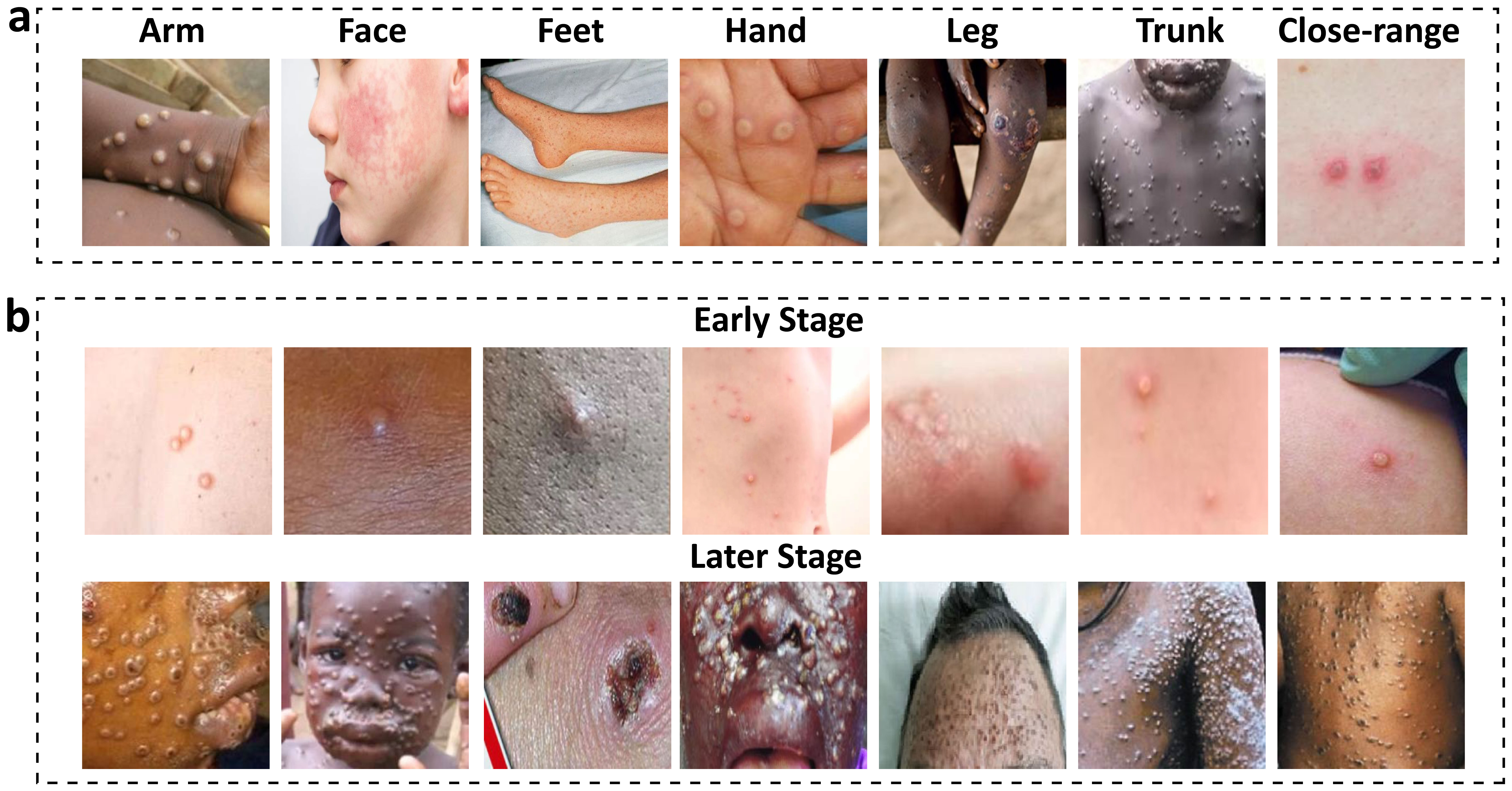}
    \caption{a: The sample of Monkeypox images from different regions in human body. b: The sample of Early-stage Monkeypox images and Later-stage Monkeypox images.}
    \label{fig9}
\end{figure}
Next, we examined seven portions and two stages of these images using Fast-MpoxNet(Aug and ImageNet). According to the World Health Organization, monkeypox is prone to occur on the face and hands of the human body \cite{48}. Additionally, detecting every early-stage monkeypox patient as much as possible is critical to controlling virus transmission. Therefore, the detection rates of these human parts are particularly important. A detailed display of the recall values for each portion and each stage have been tabulated in the Table 9. It can be seen that the average recall values of Fast-MpoxNet on early, face, and hand reached 93.65\%(±1.43), 96.95\%(±2.37), and 97.40\%(±1.45), respectively. This demonstrates the strong diagnostic ability of Fast-MpoxNet. In addition, the recall values in other parts also exceeded 90\%. These excellent recall values suggest that Fast-MpoxNet is capable of being applied in various settings.
\begin{table}[]
\caption{The Recall of original Monkeypox images from different regions in human body and  different stages.}
\resizebox{1\columnwidth}{!}{
\begin{tabular}{@{}llllllllll@{}}
\toprule
                 & \textbf{Early} & \textbf{Later} & \textbf{Face}  & \textbf{Hand}  & \textbf{Arm} & \textbf{Close-range} & \textbf{Foot} & \textbf{Leg} & \textbf{Trunk} \\ \midrule
\textbf{Fold1}   & 94.12\%        & 99.52\%        & 100\%          & 96.97\%        & 100\%        & 98.39\%              & 100\%         & 100\%        & 97.33\%        \\
\textbf{Fold2}   & 92.70\%        & 97.73\%        & 97.33\%        & 96.95\%        & 95\%         & 97.28\%              & 100\%         & 100\%        & 91.46\%        \\
\textbf{Fold3}   & 96.21\%        & 98.00\%        & 93.48\%        & 98.61\%        & 100\%        & 97.93\%              & 100\%         & 100\%        & 98.39\%        \\
\textbf{Fold4}   & 93.08\%        & 98.76\%        & 98.78\%        & 99.32\%        & 100\%        & 95.63\%              & 100\%         & 100\%        & 93.44\%        \\
\textbf{Fold5}   & 92.14\%        & 97.22\%        & 95.18\%        & 95.16\%        & 100\%        & 95.16\%              & 100\%         & 100\%        & 96.10\%        \\
\textbf{Average} & 93.65\%(±1.43) & 98.25\%(±0.81) & 96.95\%(±2.37) & 97.40\%(±1.45) & 99\%(±2.00)  & 96.88\%(±1.27)       & 100\%(±0.00)  & 100\%(±0.00) & 95.34\%(±2.55) \\ \bottomrule
\end{tabular}}
\end{table}
\subsection{Fast-MpoxNet Interpretability}
Deep neural networks are well-known for their excellent performance in various artificial intelligence tasks. However, due to their excessively parameterized black-box nature, the prediction results of deep models are often difficult to interpret \cite{49}.  In the medical field, the inability to interpret the prediction results of deep learning models can lead to difficulties for physicians in determining whether to follow the recommendations of the model for treatment. If the prediction results of the model are inexplicable, misdiagnosis and erroneous decisions may occur, which can result in serious health impacts for patients. Therefore, the interpretability of deep learning models is of paramount importance, as it can assist in comprehending the prediction results of the model, enhancing the model's credibility, reducing the risk of misdiagnosis and erroneous decisions, and ensuring the model's prediction results comply with ethical and legal regulations. 
\begin{figure}[h]
    \centering
    \includegraphics[width=1\textwidth]{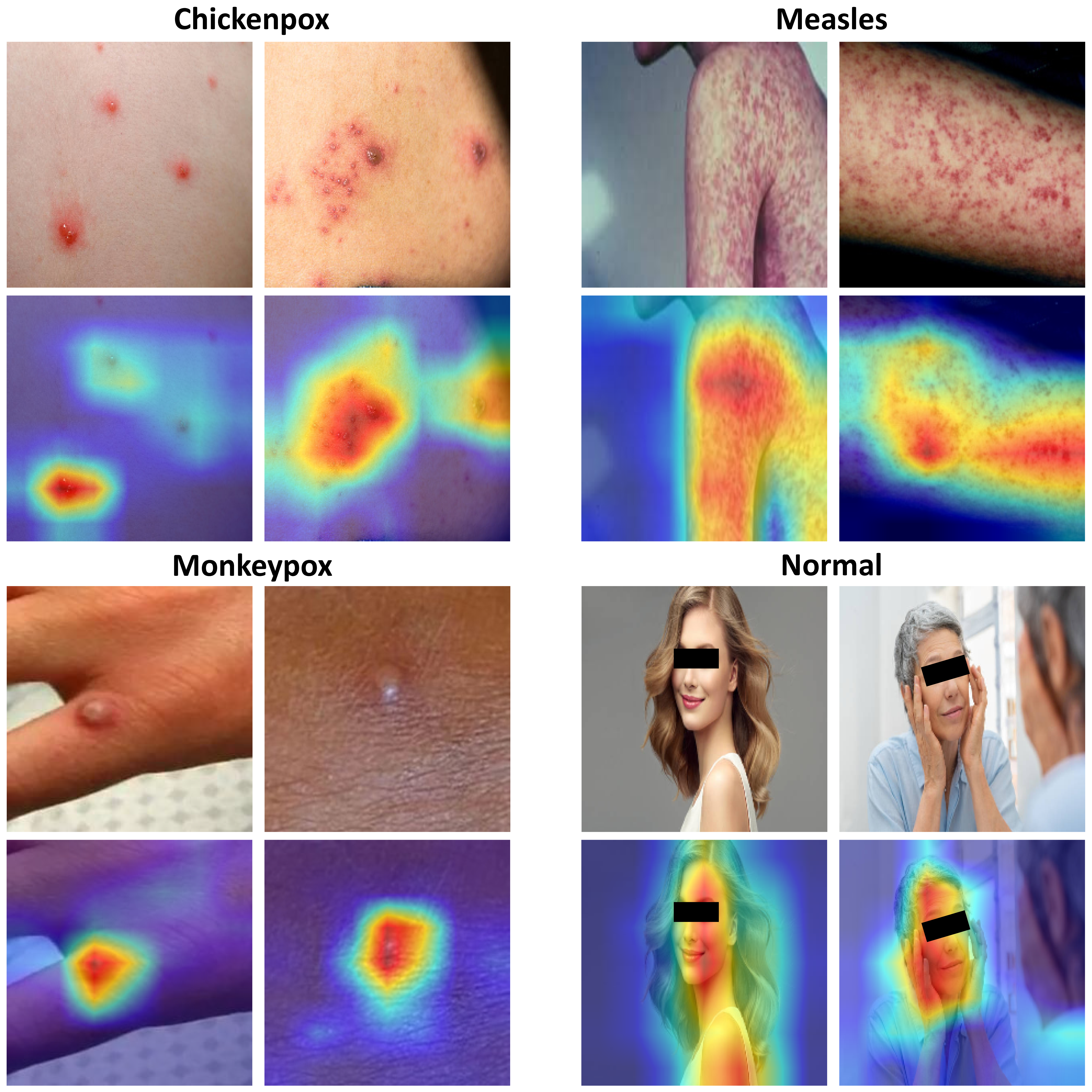}
    \caption{The heatmap generated by Grad-CAM.}
    \label{fig10}
\end{figure}
At present, there are numerous approaches available to explain the prediction results of deep learning models. In our work, we used Gradient-weighted Class Activation Mapping (Grad-CAM) proposed by Selvaraju et al. in 2016 \cite{14}. Grad-CAM (Gradient-weighted Class Activation Mapping) is a widely used technique for visualizing the activation patterns of deep neural networks in response to input images. It provides a class-discriminative localization map by producing a weighted combination of the feature maps in the final convolutional layer of a neural network, using the gradients of the predicted class with respect to the feature maps. The main advantage of Grad-CAM over other visualization methods is that it can be applied to a wide range of CNN architectures without requiring any modification to the network itself.

Additionally, it has been shown to produce more accurate and robust visualizations than previous methods. The Grad-CAM results of Fast-MpoxNet trained by Aug Dataset for each category is shown in Figure 10. By employing the Grad-CAM technique, it can be observed that Fast-MpoxNet is capable of effectively focusing on the key regions of different images.

\subsection{Deploying Fast-MpoxNet into Mpox-AISM V2}
Following an exhaustive assessment encompassing diverse metrics, graded evaluations, and model interpretability, we discerned the latent potential of Fast-MpoxNet to metamorphose into a specialized intelligent system. Consequently, we embarked on the retraining of Fast-MpoxNet using images sourced from the Aug dataset. Following this, we harnessed JAVA and Python to develop PC and mobile applications, christened as Mpox-AISM V2 (as depicted in Figure 11). The PC version (illustrated in Figure 11b-c) is underpinned by OpenCV and is adept at autonomously detecting and diagnosing skin lesions, particularly on the face and hands—regions frequently afflicted by monkeypox rashes—via real-time video. Owing to Fast-MpoxNet's modest hardware requisites and impressive average FPS, the PC version seamlessly integrates with a gamut of computing devices. This versatility empowers medical staff to orchestrate expansive screenings in high-risk, densely populated locales, thereby facilitating the timely identification of monkeypox cases and stymieing the virus's propagation. The mobile version (Figure 11a) is tailored for the layperson. Users can effortlessly capture images of suspect skin lesions, upload them to Mpox-AISM V2, and promptly receive diagnostic feedback.
\begin{figure}[ht]
    \centering
    \includegraphics[width=0.5\textwidth]{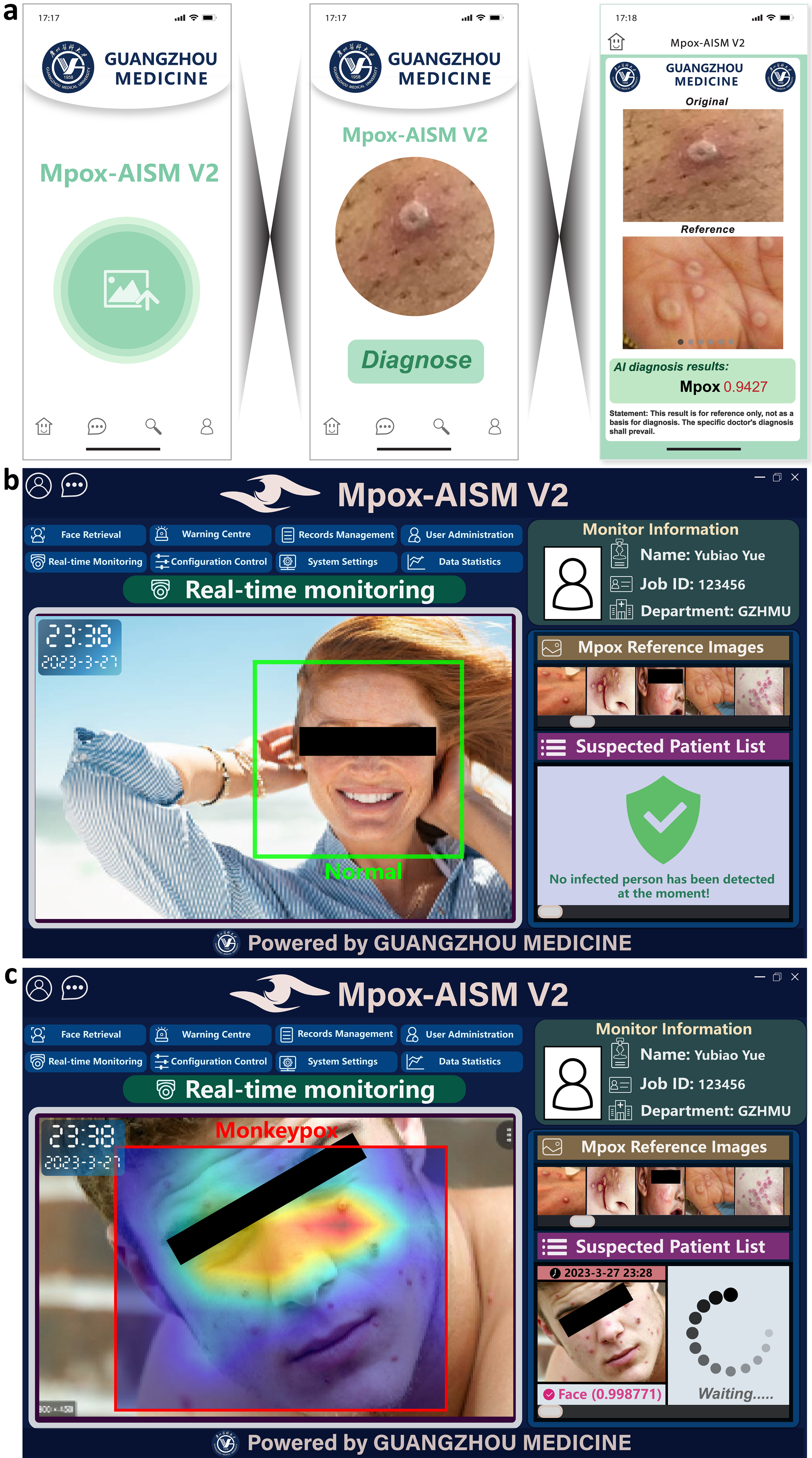}
    \caption{a: The application page of Mobile Mpox-AISM V2.
b: The application page of PC Mpox-AISM V2. When normal skin is detected, the detection box in the video window is green, and a safety sign appears in the bottom right corner of the application interface. c: The application page of PC Mpox-AISM V2. When a suspicious monkeypox patient is detected, the detection box in the video window is red and displays the corresponding heatmap. The suspicious patient list on the application interface will take a screenshot of the detected area and display risk coefficient, and detection time.
}
    \label{fig11}
\end{figure}
For empirical validation, we deployed PC Mpox-AISM V2 on a home laptop powered by an Intel Core i5 10500 and running the Windows 11 OS. Given the integration of OpenCV's detection capabilities and Grad-CAM's heatmap functionality in the PC application, the FPS oscillated between 19 and 26. However, upon deactivating the heatmap feature, the FPS surged to a range of 43 to 48. A salient attribute of Mpox-AISM V2 is its independence from internet connectivity and high-end computational infrastructure. This positions it as an invaluable tool, especially in regions grappling with rudimentary medical amenities yet vulnerable to monkeypox outbreaks. With the increasing number of monkeypox cases in the future, Mpox-AISM V2 may stand poised to play a pivotal role in preempting and managing potential outbreaks.

\section{Discussion}
A pivotal strategy in curbing the proliferation of the monkeypox virus lies in the swift and efficient diagnosis of early-stage monkeypox cases within populous settings. So, our final objective is to devise an intelligent diagnostic application for diagnosing early-stage monkeypox anytime, anywhere. To this end, we embarked on a meticulous review of related work, distilling the pivotal attributes requisite of monkeypox diagnostic models for practical deployment. Anchored by these insights, our endeavors were channeled towards harmonizing average FPS, model parameter size, and diagnostic performance for early-stage monkeypox. ShuffleNetV2 served as our foundational architecture, which we subsequently enhanced by employing the ABLGFM, the multiple auxiliary losses enhancement strategy, DropBlock and GELU. The entire improvement process was supported by exhaustive ablation studies and various metrics and underpinned by a five-fold cross-validation methodology. The resultant new architecture, christened Fast-MpoxNet, boasted an average FPS of 68 and the parameter size of 0.27M, rendering it apt for deployment across a spectrum of devices. 

To ascertain its real-world viability, we introduced the 'Practicality Score' metric, designed to measure a model's suitability for PC-based applications.  A salient aspect of our evaluation was the graded evaluation of Fast-MpoxNet for monkeypox and Grad-CAM. Compared to the ShuffleNetV2, Fast-MpoxNet without transfer learning achieved Significant improvement on the Mpox Dataset. With the incorporation of transfer learning, its performance metrics soared. Notably, its Recalls for the faces, hands and early-stage monkeypox achieve dazzling achievements. In the real-world applications, Fast-MpoxNet's performance is SOTA across all comparison models. Building on this foundation, we developed Mpox-AISM V2, an advanced diagnostic application powered by Fast-MpoxNet. This application, available in both PC and mobile iterations, caters to the diagnosis needs of medical staff and the public. While the PC variant empowers medical staff to conduct real-time monitoring in high-risk, densely populated areas, the mobile version offers a user-friendly interface for laypersons, facilitating self-diagnosis. Our detailed experimental results prove Mpox-AISM V2's potential as a frontline tool for early-stage monkeypox diagnosis in real-world settings and can position it as a potent asset in future outbreak mitigation works.
 
However, our study is not without its limitations. Despite our extensive data augmentation efforts, synthetically augmented images cannot wholly replicate the diversity inherent to genuine monkeypox images. Consequently, when confronted with data that deviates markedly from the dataset's feature distribution, Mpox-AISM V2's performance might be sub-optimal. There is also a need for rigorous clinical validation of MpoxAISM from a scientific perspective. Moreover, there's an imperative to expand the model's diagnostic repertoire to encompass a broader array of skin afflictions, offering users a more granular diagnostic insight. Recently, many regions in China have seen many cases of monkeypox, which drives us to these areas for data collection. Concurrently, we will iterative refinements to Fast-MpoxNet, enhancing its capability to discern more dermatological conditions similar to monkeypox.

In summation, juxtaposed against extant research, our endeavor started from the emphasis for model practicability and culminated in the successful development of an intelligent diagnostic system for both PC and mobile platforms, facilitating real-time diagnosis for early-stage monkeypox. Our seminal contributions encompass the delineation of monkeypox model characteristics, the design of Fast-MpoxNet, the introduction of a novel metric tailored to gauge the efficacy of real-time disease diagnostic applications and development of Mpox-AISM V2. Fast-MpoxNet adeptly harmonizes average FPS, model parameter size, and diagnostic performance for early-stage monkeypox. Notably, it exhibits excellent recall for monkeypox, especially for early-stage, face and hand, and boasts an impressive Practicality score. All these findings underpin Mpox-AISM V2 application and underscore its potential as an instrumental asset in preempting and navigating future monkeypox outbreaks.

\section*{Acknowledgments}
This work was financially supported by the Science and Technology Planning Project of Guangzhou (No. 006259497026), the Young Creative Talents of Department Education of Guangdong (Natural Science, No. 2019KQNCX067), the National Natural Science Foundation of China (Grant No. 52172083), Guangdong Natural Science Foundation (Grant No. 2019A030310444), International Science \& Technology Cooperation Program of Guangdong (Grant No. 2021A0505030078), Guangzhou Education Bureau University Research Project: Youth Talent Research Project (Grant No. 202235325).
\bibliographystyle{unsrt}  
\bibliography{references}

\end{document}